\pdfoutput=1

\documentclass[11pt]{article}

\usepackage[]{acl}

\usepackage{times}
\usepackage{latexsym}
\usepackage{arydshln}
\usepackage{rotating}

\usepackage[T1]{fontenc}

\usepackage[utf8]{inputenc}

\usepackage{microtype}
\usepackage{graphicx}
\usepackage{enumitem}
\usepackage{xcolor}

\usepackage[absolute,overlay]{textpos}
\usepackage{amsmath}
\usepackage{amssymb}

\title{Do BERTs Learn to Use Browser User Interface? \\
Exploring Multi-Step Tasks with Unified Vision-and-Language BERTs
}

\author{Taichi Iki \and Akiko Aizawa \\
National Institute of Informatics, Chiyoda-ku, Tokyo, Japan \\
Graduate University for Advanced Studies, Hayama, Kanagawa, Japan \\
\texttt{\{iki,aizawa\}@nii.ac.jp}
}

\begin{document}

\begin{textblock*}{\textwidth}(2.5cm, 2cm)
\centering
{\large{Work in Progress}}
\end{textblock*}

\setlength{\abovedisplayskip}{3pt}
\setlength{\belowdisplayskip}{3pt}
\setlength{\intextsep}{5pt}
\setlength{\parskip}{0pt}

\maketitle
\begin{abstract}
\textcolor{black}{Pre-trained Transformers are good foundations for unified multi-task models owing to their task-agnostic representation.
Pre-trained Transformers are often combined with text-to-text framework to execute multiple tasks by a single model. 
Performing a task through a graphical user interface~(GUI) is another candidate to accommodate various tasks, including multi-step tasks with vision and language inputs. 
However, few papers combine pre-trained Transformers with performing through GUI.}
To fill this gap, we explore a framework in which a model performs a task by manipulating the GUI implemented with web pages in multiple steps.
We develop task pages with and without page transitions and propose a BERT extension for the framework.
We jointly trained our BERT extension with those task pages, and made the following observations.
(1) The model learned to use both task pages with and without page transition. 
\textcolor{black}{(2) In four out of five tasks without page transitions, the model performs greater than 75\% of the performance of the original BERT, which does not use browsers.}
(3) The model did not generalize effectively on unseen tasks.
\textcolor{black}{These results suggest that we can fine-tune BERTs to multi-step tasks through GUIs, and there is room for improvement in their generalizability. 
Code will be available online\footnote{\textcolor{black}{Repository:  \url{https://github.com/Alab-NII/bertbui_pub/}}}.}
\end{abstract}

\section{Introduction}\label{sec:introduction}

\begin{figure}
\centering
\includegraphics[width=1.0\linewidth]{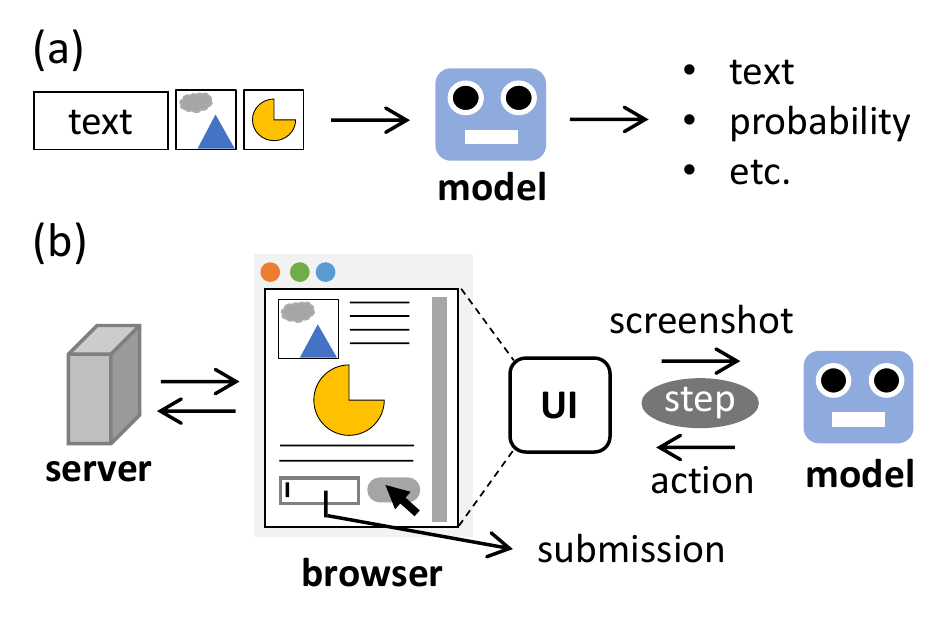}
\vspace{-8mm}
\caption{\label{figure:concept}
\textcolor{black}{
Framework comparison.
(a) Typical frameworks assume single-step tasks in which a model takes a sequence of text and images to generate an output.
(b) In our WoB-like framework, we make a task as web pages, which allow structured contents, hyperlinks and scripts.
The page design decides how to \emph{submit} an answer (e.g., choose a button or input text).
A model completes a task in \emph{multiple steps} using the browser user interface~(BUI).
The model take a screenshot to output an action for each step. (e.g., click or keystroke).}}
\end{figure}

Prior studies have attempted to unify models for processing natural language to facilitate the transfer of learned knowledge by reducing task-specific structures.
For example, \citet{radford2018improving, devlin2019bert} suggest that language models with a generic structure, Transformer~\cite{vaswani2017attention}, are effective.
\citet{raffel2020exploring} proposed a text-to-text framework which converts tasks into a problem where a model receives and generates text.
\citet{cho2021unifying} extended the input of the text-to-text framework to accommodate images.

However, existing research on unified models remains limited.
First, although the models proposed by \citet{cho2021unifying} use a linear sequence of text and several images as input, their models are not designed to handle input with a layout.
Second, existing unified models assume single-step tasks. 
Task-specific design still must be completed when applying these models to compound tasks, such as reading a single document and subsequently searching for missing information.
The latter challenge is more difficult to address because methods for using a transformer in multiple-step tasks, have not yet been fully established.
Although transformer-based models have been successful in many language-related tasks such as language understanding~\cite{wang2019glue}, question answering~\cite{rajpurkar2016squad}, visual question answering~\cite{antol2015vqa}, and referring expression comprehension~\cite{kazemzadeh2014referitgame}, nevertheless, these are single-step tasks.

\textcolor{black}{In this study, we investigate the following question to explore the possibility of performing tasks:}
\emph{Can language models complete tasks through graphical user interfaces expressed as visual input?}
\textcolor{black}{
For graphical user interface manipulation, \citet{shi2017world} introduced the World of Bits~(WoB) platform, in which tasks are expressed as web pages, and models complete those tasks via browser UIs.
We follow their work and apply language models to a similar task framework~(Figure~\ref{figure:concept}).} 

We formulate the interaction between a browser and a model (\S~\ref{sec:target_bui}), and create task pages based on the existing datasets, including GLUE~\cite{wang2019glue}, SQuAD~\cite{rajpurkar2016squad} and VQA~\cite{antol2015vqa} (\S~\ref{sec:tasks}).
Our tasks include not only single-page but also multi-page tasks that require page jumps to diversify the goal of actions.
We introduce a BERT~\cite{devlin2019bert} extension with a simple memory mechanism and pre-training for actions (\S~\ref{sec:baseline}).
In our experiments, we train our model in a multi-task setting.
We validate whether our model can learn in the framework and compare it with other models based on the same BERT.
We show that our pre-training and memory mechanisms are effective and analyze the models' ability to solve unseen tasks (\S~\ref{sec:experiments}).

\vspace{0.25em}\noindent\textbf{Our contributions:}
\begin{itemize}[leftmargin=*]
\setlength{\itemsep}{-3pt}
\item \vspace{-4pt} \textcolor{black}{We construct a task set to evaluate how well a language model is transferred to the GUI manipulation retaining its linguistic knowledge.}
\item We introduced a BERT extension and demonstrate its ability to learn diverse tasks (GLUE, SQuAD, VQA, and multi-page tasks) jointly.
\end{itemize}

\section{Related Work}\label{sec:related_work}

\subsection{Execution Style of Unified Models}

Unified Models aim to reduce task-specific structures to promote learning different tasks jointly such that learned knowledge can be shared between tasks\footnote{While it is a kind of multi-task learning~\cite{caruana1997multitask, ruder2017overview}, it often does not have the central tasks.}.
After the success of transformer-based language models (LMs)~\cite{devlin2019bert, radford2019language} and their visual extensions~\cite{NEURIPS2019_c74d97b0, li2019visualbert, tan-bansal-2019-lxmert, chen2020uniter, Su2020VL-BERT:}, unified models with transformers have received significant attention. 

We can categorize unified transformers in terms of task execution: {\it task-specific head} and {\it text generation} styles.
The task-specific head style shares most model weights between tasks and provides a head for each task.
ViLBERT-MT~\cite{lu202012} and UniT~\cite{hu2021unit} use this style.
The text generation style employs text generation to bridge the differences in output between tasks.
GPT-2~\cite{radford2019language} and GPT-3~\cite{brown2020language} show that large pre-trained models can multitask in the text region by changing the prompt.
T5~\cite{raffel2020exploring} and VL-T5~\cite{cho2021unifying}, which extends T5 to vision, also employ this style.
The text generation style can be applied to, in principle, all tasks that can be expressed in a text-to-text format.
\textcolor{black}{
Our WoB-like framework is also highly flexible.}
A model manipulates web pages that define a task via general actions.
As a result, it extends the tasks to what can be rendered in a browser screen while keeping the model structure.
We refer to this style as the BUI action style.

\subsection{Vision-and-Language Tasks}
\textcolor{black}{\noindent{\bf World of Bit platform}~\cite{shi2017world} is a reinforcement learning environment for web tasks, which is most relevant to our work. 
The MiniWoB benchmark and its extension MiniWoB++~\cite{liu2018reinforcement} on the platform are widely used to study models for web tasks.
They consist of simple tasks such as clicking a button, typing text or more complex tasks such as booking a flight.
Although prior work on those benchmarks spans architecture/policy~\cite{liu2018reinforcement, gur2018learning, jia2018domqnet, humphreys2022data}, parsing~\cite{srivastava-etal-2020-learning}, and environment generation~\cite{gur2021environment}, Few studies have investigated the use of LM's linguistic knowledge in the task procedures.
}

\noindent{\bf Document AI} is a technique for automatically reading, understanding, and analyzing documents~\cite{cui2021document}.
Our work relates to studies on HTML documents.
\citet{tanaka2021visualmrc, chen2021websrc} proposed reading comprehension datasets on web pages.
\citet{wu2021lampret, li2021markuplm} proposed pretrained models for HTML documents. 
Although documents are processed differently (such as using screenshots or incorporating hierarchy of the elements), prior studies were concerned with a visually rich layout.
Our focus is on the interaction between models and the documents.

\noindent{\bf UI modeling} is an emerging topic, and \citet{baiuibert, he2021actionbert} have pre-trained UI models for mobile devices to obtain better representations for the UI in terms of understanding tasks, such as predicting the application type or retrieving similar UI components.
\citet{li2021vut} proposed a multi-task UI model that can answer questions about the UI.
While the questions include commands e.g., 'Go to the next screen', they are limited to single-step commands. 
By contrast, our models use UIs by recurrently generating actions.

\noindent{\bf Vision-and-language 
navigation~(VLN)}~\cite{anderson2018vision, das2018embodied, shridhar2020alfred} studies models that follow instructions in a physical space, such as room.
VLN tasks have progressed in action generation with V\&L models.
Recent studies used pre-trained LMs to encode instructions~\cite{li2019robust, majumdar2020improving, hong2021vln, Qi_2021_ICCV}.
However, the visual input rarely contains long text because the target is a physical space.
Combination of views with a long text and actions remains a challenge.

\section{Task Formulation with Browser UI}\label{sec:target_bui}
\textcolor{black}{In this study, the term browser refers to software for accessing web pages.
A browser renders web pages, navigates to a new page when a hyperlink is clicked, and executes the scripts on a page internally.}

Our formulation focuses on browsers that run on personal computers\footnote{
Firefox (\url{https://www.mozilla.org/en-US/firefox/}) was adopted as the browser and Selenium (\url{https://www.selenium.dev/}), which is an automation tool for browser operations, was used to apply the model’s actions to the browser.
}.
We assume that the browser input devices are a mouse and keyboard, and that the browser provides a screenshot. 
At each step, the model partially observes the state of web pages from a screenshot to output an action.
\textcolor{black}{The cursor position is drawn as a dot in the screenshot.}
We apply the action to the browser and waited for a period of time ($\sim 500$ms)\footnote{
An internal server was used.
Accessing external servers could require additional time.
} for the browser to complete internal computation (e.g., rendering, navigating).
Subsequently, we take the next screenshot.
In conclusion, suppose a screenshot of the visible area of a page~$s_i$ and model's action~$a_i$ at step $i$, then the model predicts $a_i$ from $s_i$:
\textcolor{black}{
\begin{align*}
a_i &= {\rm Model} (s_0, .., s_i, a_0, ..., a_{i-1}), \\
s_{i+1} &= {\rm Browser} (a_i).
\end{align*}
Since our formulation does not assume Markovian, we consider all previous screenshots and actions in the above equation. Note that models are not required to use all of them.
}

\vspace{0.25em}\noindent\textbf{Fixed-size screenshot.}
In lieu of inputting a whole page by using a screenshot with variable size or scale, we use fixed-size screenshots and give the models actions to move their visible area.
Such actions are suitable for pages that dynamically load additional parts and avoid unexpected long inputs. 

\begin{table}
\small
\centering
\scalebox{0.97}{
\begin{tabular}{lll}
\hline
{\bf scope} & {\bf action name} & {\bf description} \\
\hline
mouse & $\rm MOVETO (x, y)$ & move the cursor to (x, y) \\
mouse & $\rm CLICK$ & right click. \\
key & $\rm TOKEN (word)$ & type characters in a word \\
key & $\rm SPACE$ & type space key \\
key & $\rm BACKSPACE$ & type backspace key \\
key & $\rm ENTER$ & type enter key \\
view & $\rm LEFT$ & move the view to the left \\
view & $\rm RIGHT$ & move the view to the right \\
view & $\rm UP$ & move the view to the up \\
view & $\rm DOWN$ & move the view to the down \\
\hline
\end{tabular}
}
\vspace{-2mm}
\caption{
Defined actions.
$\rm MOVETO$ and $\rm TOKEN$ take the arguments specified in the parentheses.
\label{figure:atomic_actions}
}
\end{table}

\vspace{0.25em}\noindent\textbf{Actions.}
Table~\ref{figure:atomic_actions} presents the actions defined.
The actions cover using a mouse, keystrokes and moving the visible area.
The unit of keystrokes is the model's vocabulary.
A model selects one action for each step.
Thus, if a task requires inputting a sentence to a text box, the model will move the cursor to a text box ($\rm MOVETO$), click it ($\rm CLICK$), and enter tokens ($\rm TOKEN$).

\vspace{0.25em}\noindent\textcolor{black}{\textbf{Difference between WoB.}
WoB uses screenshots, Document Object Models (DOM) of HTML pages for input, while our formulation uses only screenshots.
Although the works~\cite{liu2018reinforcement, gur2018learning, jia2018domqnet} following WoB use DOM-specific actions such as click(element), we use low-level actions following the original WoB.
In addition, our action formulation differs from the one of WoB in the following points: 
first, our formulation uses sub-word from language models instead of characters as the unit to type text; second, it does not use special keys such as the Ctrl key; third, it does not use drag, drop or left-click for mouse actions. Note that the last two restrictions are only for simplicity.
}

\section{Task Pages}\label{sec:tasks}
\begin{figure*}
\centering
\includegraphics[width=0.90\linewidth]{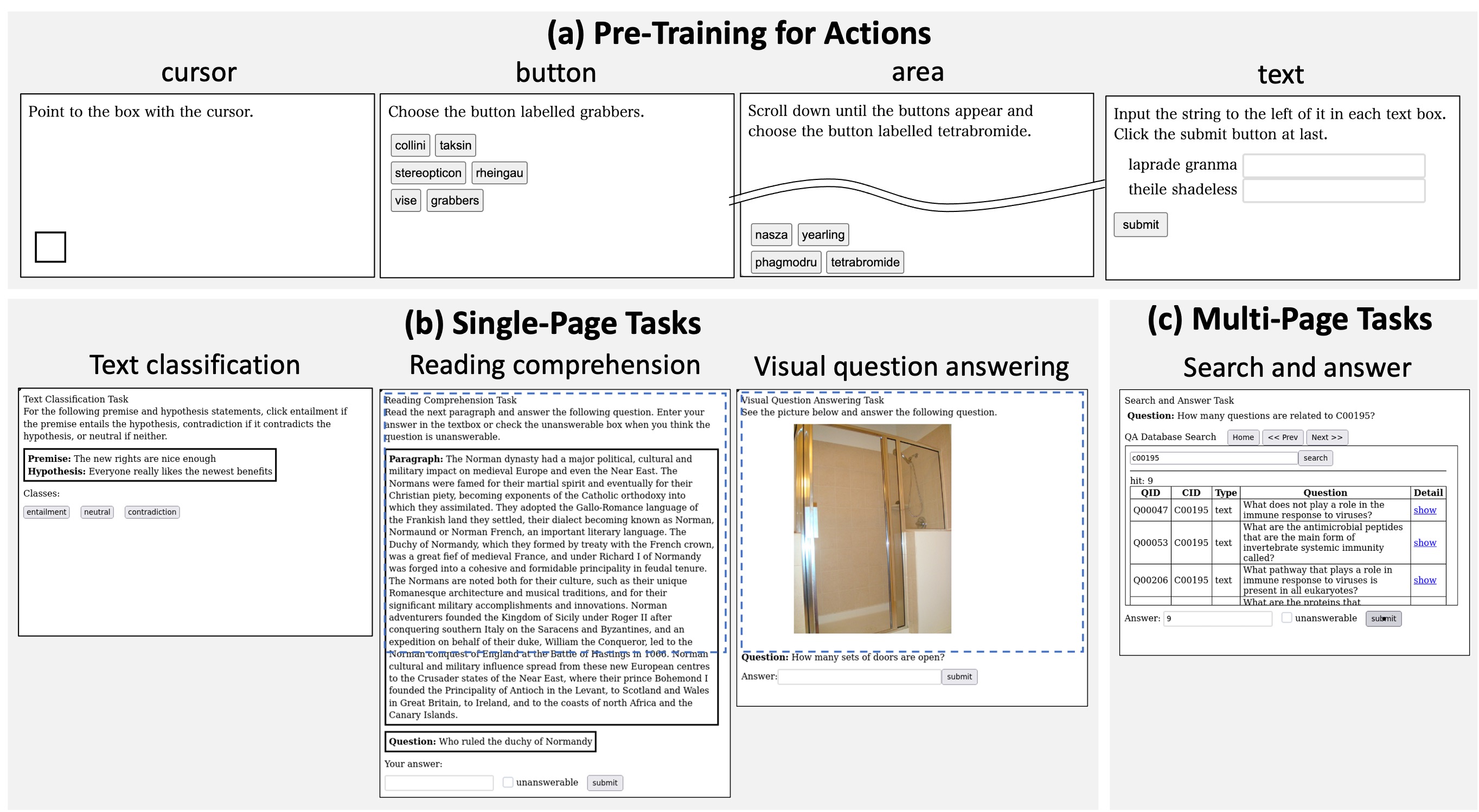}
\vspace{-3mm}
\caption{\label{figure:tasks}
\textcolor{black}{Three types of task page and the examples.
(a) Pre-training for actions.
In the \emph{area} task, a blank space exists between the instructions and the buttons so that a model needs to scroll until the buttons are visible.
(b) Single-page tasks. 
The rectangles outlined by the blue dotted lines represent the initial visible area.
(c) Multi-page tasks. 
Models can make page transitions within the child frames embedded in the outer page.}}
\vspace{-3mm}
\end{figure*}

This section describes the web pages we created for tasks.
Although there are no restrictions on the layout of the pages, we used layouts that have an instruction, a main content, and an answer form for simplicity.
We assumed that a task example has a single answer.
Figure~\ref{figure:tasks} summarizes task pages we made.
This section describes the types of task page and how to obtain gold actions for training.

\subsection{Types of Task Page}

\noindent\paragraph{(a) Pre-Training for Actions~(PTA).} \label{sect_pta}

Prior knowledge of interface usage, such as the use of clickable buttons, could assist more efficient learning of tasks in the BUI by avoiding situations where models learn such knowledge and reasoning (e.g., reading comprehension) simultaneously.
We introduced pre-training for actions: a set of small tasks that focus on moving the \emph{cursor}, clicking a \emph{button}, moving the visible \emph{area}, and inputting \emph{text}.
As shown in Figure~\ref{figure:tasks}, in PTA tasks, the instructions are written at the top of the screen, and the model succeeds if it follows the instruction.
We generated task instances using templates~(in Appendix~\ref{app_pta_templates}).

\noindent\paragraph{(b) Single-page tasks.} 
To evaluate to what extent models can solve traditional tasks in BUI, we created tasks of this type based on existing datasets.
We used GLUE~\cite{wang2019glue} and SQuAD~\cite{rajpurkar2016squad, rajpurkar2018know} for natural language understanding and VQA~\cite{antol2015vqa} for visual grounding.
\textcolor{black}{Task pages of this type involve scrolling pages and submitting answers.}
We chose answer forms that matched the format of those datasets.
We used buttons for GLUE (classification), and a text box for SQuAD and VQA (question answering).
The condition for success is to submit the correct answer of the original datasets.

\noindent\paragraph{(c) Multi-page tasks.}
\begin{figure}
\centering
\includegraphics[width=0.70\linewidth]{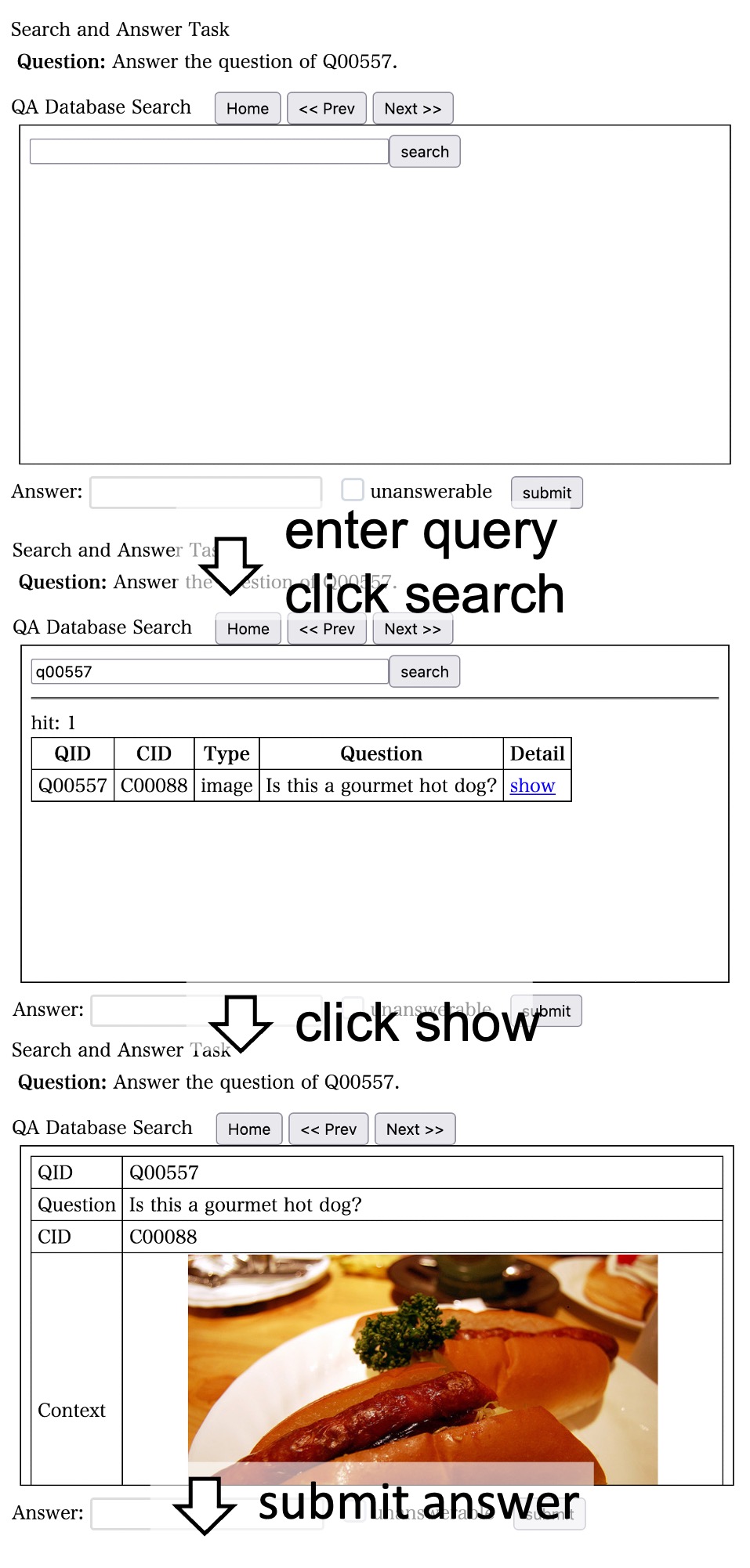}
\vspace{-4mm}
\caption{\label{figure:sa_example}
\textcolor{black}{Example pages for the SA-A task.
All the SA tasks share the page design.
The task pages include an initial page, search result, and detail page.
The result changes depending on the query.
Models are required to jump between those pages to answer the question.}}
\end{figure}

This type introduces page transitions to focus more on procedural tasks that BUI enables.
We designed Search and Answer~(SA) task~(Figure~\ref{figure:sa_example}).
For the task, we made databases on question answering tasks by sampling the contexts (paragraphs or images) and questions from SQuAD and VQA.
We assigned unique ids to the contexts and questions.
Task pages of SA are linked to one of those databases that can be queried with the search UI.
The goal of the tasks is to answer a question about the database using the search UI.
We prepared four groups to verify whether the models can handle different questions:
\begin{itemize}[leftmargin=*]
\setlength{\itemsep}{-3pt}
\item \vspace{-5pt} \textbf{SA-H}: \textcolor{black}{\emph{How many questions are related to CID?}} \\
requires querying a given Context ID (CID) and answering the number of Hits.
\item \noindent\textbf{SA-Q}: \textcolor{black}{\emph{What is the question of QID?}} \\
requires identification of the question corresponding to a given Question ID~(QID). 
\item \noindent\textbf{SA-QID}: \textcolor{black}{\emph{What is the QID of QUESTION?}}\\
requires identification of the QID corresponding to a given question. 
\item \noindent\textbf{SA-A}: \textcolor{black}{\emph{Answer the question of QID.}} \\
requires answering the question corresponding to a given QID. 
\end{itemize}
\vspace{-3pt} While SA-H, -Q, and -QID can be answered directly from the search results, SA-A requires models to display a detailed page to produce the answers.
Appendix~\ref{app:sa} provides further detail.

\begin{figure*}
\centering
\includegraphics[width=0.80\linewidth]{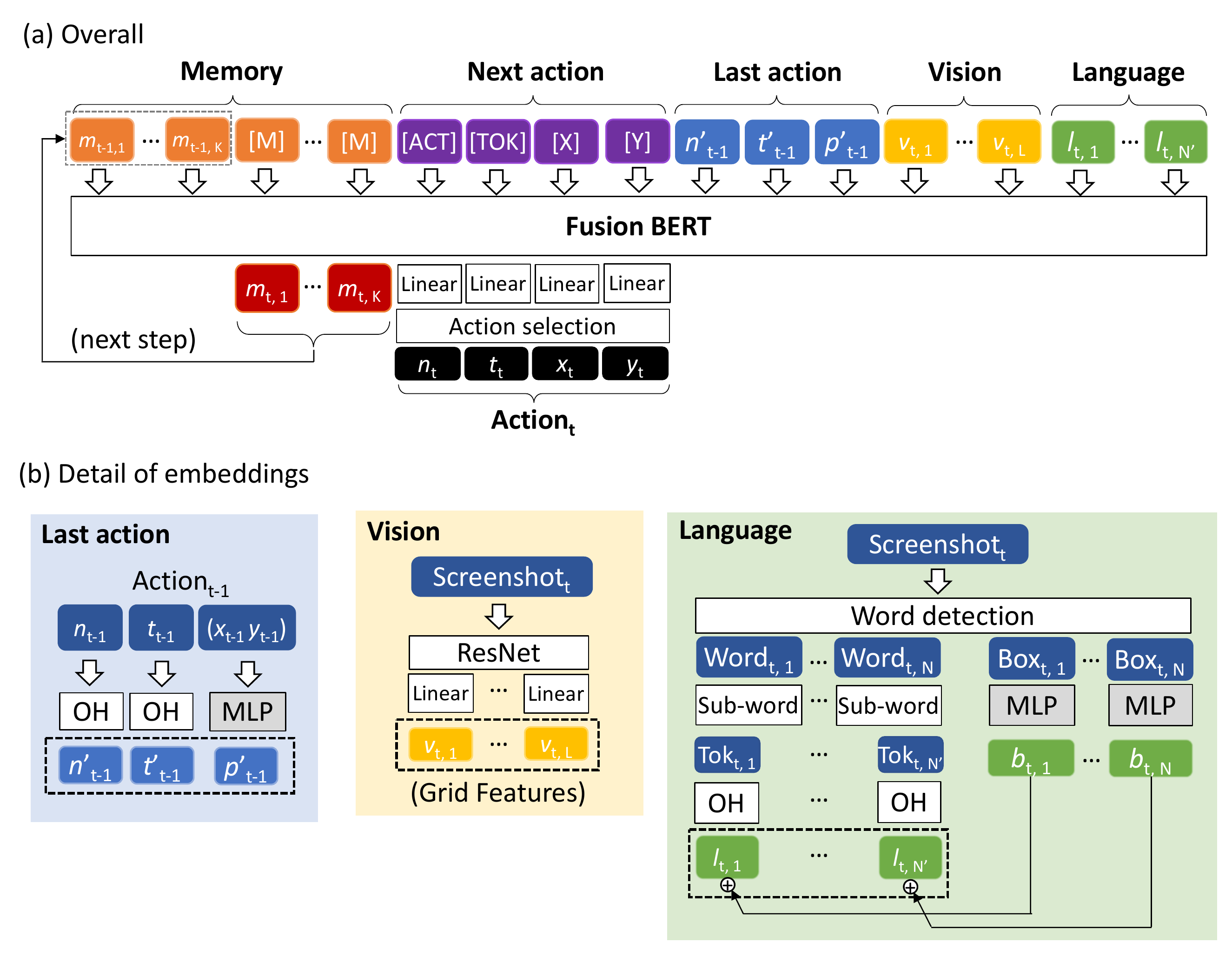}
\vspace{-4mm}
\caption{ \label{figure:model_overview}
\textcolor{black}{Overview of BUI-BERT.
This figure shows how the model predicts an action for the step $t$. 
(a) The model consists of a pre-trained BERT (Fusion BERT), which takes vision, language, memory, and some auxiliary tokens to output the next action.
Fusion BERT adds position and segment embeddings in the same way as the original BERT (omitted in the view).
(b) The model obtains action~(three tokens), vision~($L$ tokens) and language embeddings~($N'$ tokens from $N$ words) with deep neural networks.
}}
\end{figure*}

\subsection{Creating Gold Sequences}
Supervised learning was used to train the proposed BUI model because the probability of completing a task by randomly acting on a page appears small. 
To record gold sequences of action-screenshot pairs, we manually created rules for each task and manipulated task pages loaded in a browser following the rules.
We designed the rules to identify the contents on a task page once and to take actions to submit answers.

Individual rules are listed as follows.
Note that each rule breaks down further into the actions.
\begin{itemize}[leftmargin=*]
\setlength{\itemsep}{-2pt}
\item \vspace{-4pt} \noindent\textbf{GLUE.} 
Scroll down until the buttons appear (to view all contents).
Click the correct button.
\item \noindent\textbf{SQuAD and VQA.}
Scroll down until the submit form appears.
If the question is unanswerable, check unanswerable.
Otherwise, type the answer in the text box.
Click the submit button.
\item \noindent\textbf{SA-H.}
Query the CID in an instruction.
Submit the number of hit records.
\item \noindent\textbf{SA-Q.}
Query the QID in an instruction.
Scroll down until the record related to the query appears.
Submit the corresponding question.
\item \noindent\textbf{SA-QID.}
We swapped the question and QID in the SA-Q rule.
We used only the first three tokens of the question to reduce the number of actions.
\item \noindent\textbf{SA-A.}
Query QID in an instruction.
Click the 'show' link to display the context.
View the entire context to produce the answer for the question corresponding to the QID.
Submit the answer.
\end{itemize}

\section{BUI-BERT}\label{sec:baseline}

This section describes how to extend the pre-trained BERT$_{\rm small}$ to manipulate the browser UI.
A small language model was used instead of pre-trained professional models with standard size (e.g., LayoutLM) owing to the multiple long sequences required by the BUI setup.
As illustrated in Figure~\ref{figure:model_overview}, vision, language, memory, and some auxiliary tokens are fused with fusion BERT to obtain the next action.
We initialized the weight of fusion BERT based on the weight of the pre-trained BERT and pre-trained the model using the PTA tasks.

\subsection{Vision Input}

We used grid features from a pre-trained convolutional neural network similar to \citet{huang2020pixel}, considering the speed and amount of data.
We encoded the screenshot at each step with a frozen pre-trained ResNet~\cite{he2016deep}\footnote{Pre-trained ResNet18 bundled with PyTorch Vision.} followed by a trainable fully connected layer.

\subsection{Language Input}

To treat words as a separate modality, we detected words from a screenshot and broke down the words into sub-words using a BERT uncased tokenizer\footnote{\textcolor{black}{We used the BertWordPieceTokenizer of Huggingface's tokenizers~(\url{https://github.com/huggingface/tokenizers}) with the vocabulary of pre-trained BERTs.}}.
We emulated a word-based OCR by inserting span tags between words in the HTML pages.
We present a detection example in Appendix~\ref{app_detection}.
While this emulation works in text content and labels on the buttons, it does not capture text in text boxes.
Despite this limitation, we use this emulation to reduce the computation time and avoid the necessity to consider detection errors.
\textcolor{black}{
Note that models can recognize their input through the last action tokens.
}

\vspace{0.25em}\noindent\textbf{Location embedding.}
We added location embedding to each sub-word embeddings to indicate the location on a screenshot.
The word bounding box\footnote{
(center x, center y, width, and height).
All elements were normalized by the width or height of a screenshot.
} was encoded by a trainable MLP.
Sub-words in a word share the location embedding of the word.

\subsection{Memory Mechanism}
\textcolor{black}{
Our model uses memory to handle previous inputs. 
Suppose that $K$ is a hyper-parameter for the memory token length (We used 64 in this paper).
}
Our memory mechanism used $2\times K$ embeddings.
The first half $K$ was copied from the previous memory output, and the second half was filled with the [M] embedding, a trainable one-hot vector.
After fusing inputs, we retained the $K$ encoded embeddings corresponding to the second half for the next step.
During training, we inputted the memory embeddings recurrently while the number of steps did not exceed the maximum~(50 in our study).

\subsection{Auxiliary Inputs}
\noindent\textbf{Last action.}
The last action is represented with the embeddings of the action name, the cursor position, and the sub-word~\footnote{For actions unrelated to the cursor position or sub-word, their embeddings were filled with zeros.}.
We used trainable one-hot vectors for the action name and the sub-word embeddings.
We encoded the cursor position using the same MLP as the word location by setting width and height to zero.

\vspace{0.25em}\noindent\textbf{Next action.}
We appended trainable one-hot vectors for [ACT], [TOK], [X], and [Y] tokens and inputted these tokens to predict the next action.

\subsection{Next Action Prediction and Loss}
\label{sec:loss}

We predicted the next action from the embeddings that the fusion BERT encoded the [ACT], [TOK], [X], and [Y] tokens.
Suppose the encoded embeddigns are $e_{\rm act}$, $e_{\rm Tok}$, $e_{\rm x}$, and $e_{\rm y}$.
We first classified the action name from $e_{\rm act}$.
We then classified the token id in Fusion BERT's vocabulary from $e_{\rm tok}$ for $\rm TOKEN$ and the pixel coordinate\footnote{x $\in$ \{1, ..., screen width\} and y $\in$ \{1, ..., screen height\}.} from $e_{\rm x}$ and $e_{\rm y}$ for $\rm MOVETO$.
All embeddings are projected to the class distributions with trainable linear layers.
During training, we used the Softmax cross-entropy loss for the action name, token, x, and y.
These were evenly added in a mini-batch:
\begin{eqnarray*}
~~~~L_{\rm mb} = \langle L_{\rm name} \rangle + \langle L_{\rm token} \rangle + \langle L_{\rm x} \rangle + \langle L_{\rm y} \rangle,
\end{eqnarray*}
where $\langle \cdot \rangle$ denotes average for non-pad labels.

\textcolor{black}{Note that although our loss function does not explicitly refer to memory usage, our results in Section~\ref{sec:pta} and \ref{sec:analysis} show that the models learn to use memory with this loss.}

\section{Experiments\footnote{Note that our results are based on a single run.}} \label{sec:experiments}
First, we trained our BUI models on the PTA tasks to pre-train the models.
Second, we trained the BUI models in the multi-task setting; thereafter, we compared the BUI models to the models with different task styles.
Finally, we analyzed the models.

\subsection{Pre-Traing for Actions}
\label{sec:pta}
\begin{table}
\small
\centering
\scalebox{1.0}{
\begin{tabular}{lcccc}
\hline
{\bf model} & {\bf cursor} & {\bf button} & {\bf area} & {\bf text} \\
\hline
BUI-BERT$_{\rm small}$ & 1.00 & 0.90 & 0.56 & 0.87 \\
BUI-BERT$_{\rm medium}$ & 1.00 & 0.62 & 0.49 & 0.88 \\
\hline
chance level & - & 0.43 & 0.42 & - \\
\hline
\end{tabular}
}
\vspace{-2mm}
\caption{\label{table:result_pta}
Exact match accuracy of BUI-BERTs on Pre-Training for Actions.
Models were trained on the four tasks jointly.
We show reciprocals of the number of buttons as chance levels of the button and area tasks.
\textcolor{black}{The ratio of the cases in which a model did not submit an answer, i.e., timeout, was lower than 0.03 for all tasks.}}
\end{table}

We trained small and medium sized BUI-BERTs\footnote{Initialized with the pre-trained BERTs from https://github.com/google-research/bert} on PTA tasks jointly with 60k training examples.
\vspace{0.25em}\noindent\textbf{Setup.}
The memory length of both models was 64.
We set the screen at 640px$\times$448px and resized the screenshots by half before inputting to ResNet18.
The maximum epoch was 50.
We tracked the validation loss at the end of each epoch and used the model with the smallest validation loss for the evaluation with the actual browser.
During evaluation, the trial was stopped and considered a failure if a model did not submit an answer within 1.5 times the number of steps in the gold sequences.
We used the ADAM optimizer~\cite{kingma2014adam} with a fixed learning rate of 5e-5 and accumulated the gradient such that the mini-batch size was 128.

\vspace{0.25em}\noindent\textbf{Results.}
Table~\ref{table:result_pta} presents the results of the PTA tasks.
Our models performed well in the cursor, button, and text tasks, \textcolor{black}{showing that we can fine-tune language models on such simple actions. 
The model scores are above the chance level on the area task, which requires models to remember the instruction to choose a correct button.
This suggests the effectiveness of our memory mechanism.
BUT-BERT$_{\rm medium}$ performs worse than BUT-BERT$_{\rm small}$ in the button and area tasks despite its larger parameter size.
The reason is not yet clear.
Considering that both scored at the same level on the text task, a possible interpretation may be that the medium model did not learn the location embedding of sub-words well and could misplaced the sub-words on the view.
}

\subsection{Main Tasks} \label{subsec:main_experiment}

\begin{table*}
\small
\centering
\begin{tabular}{lllllll}
\hline

{\bf model} & {\bf base LM} & {\bf \#params} & {\bf exec. style} & {\bf architecture} \\ 
\hline

BERT$_{\rm small}$ / +V & 
~ &
31M / 42M &
task-spec. head & 
BERT~\cite{devlin2019bert} \\ 

BERT$_{\rm small}$-s2s+V &
BERT$_{\rm small}$ &
74M &
text gen. &
Enc-dec from Pr. LMs~\cite{rothe2020leveraging} \\ 

BUI-BERT$_{\rm small}$ &
~ &
42M &
BUI action &
BUI-BERT \\ 
\hline

BUI-BERT$_{\rm medium}$ &
BERT$_{\rm medium}$ &
54M & 
BUI action &
BUI-BERT \\ 
\hline

\end{tabular}
\vspace{-2mm}
\caption{\label{table:models}
Models to be compared.
Model with +V use an image input obtained from a frozen pre-trained ResNet18.
Of the \#params, ResNet18 and its related layers account for approximately 11M.
}
\vspace{-2mm}
\end{table*}

\begin{table*}
\small
\centering
\scalebox{1.0}{
\begin{tabular}{lcc|ccccc|c}
\hline
~ & ~ & ~ & ~ & \multicolumn{3}{c}{\hspace{5mm}\textcolor{black}{non-action or single-page}} & ~ & \textcolor{black}{multi-page} \\
{\bf model} &
{\bf multi-task} &
{\bf \textcolor{black}{multi-step}} &
\begin{tabular}[c]{@{}c@{}} {\bf CoLA} \\ M \end{tabular} &
\begin{tabular}[c]{@{}c@{}} {\bf STS-B} \\ P \end{tabular} &
\begin{tabular}[c]{@{}c@{}} {\bf MNLI-m} \\ macro f1 \end{tabular} &
\begin{tabular}[c]{@{}c@{}} {\bf VQAv2} \\ acc. \end{tabular} &
\begin{tabular}[c]{@{}c@{}} {\bf SQuADv2} \\ EM \end{tabular} &
\begin{tabular}[c]{@{}c@{}} {\bf SA} \\ acc. \end{tabular} \\
\hline

BERT$_{\rm small}$/+V & no &  &
31.3 & 81.2 & 75.8 & 42.9/51.4 & 56.8 & - \\

BERT$_{\rm small}$-s2s+V & w/o SA & &
0.0 & 82.5 & 75.5 & 51.4 & 47.0 & - \\

BUI-BERT$_{\rm small}$ & all & \checkmark &
-1.0 & 71.6 & 70.4 & 48.0 & 43.6 & 75.5 \\

\hline

BUI-BERT$_{\rm med.}$ & all & \checkmark &
-1.8 & 78.3 & 75.6 & 48.8 & 49.4 & 69.7 \\

\hline

\end{tabular}
}
\vspace{-2mm}
\caption{\label{table:results_main}
\textcolor{black}{
Overall scores on the validation splits.
M and P denote Matthews' and Pearson's correlation, respectively.
}}
\end{table*}

We trained the pre-trained BUI-BERTs on CoLA~\cite{warstadt2019neural}, STS-B~\cite{cer2017semeval}, MNLI-matched~\cite{williams2018broad}, SQuADv2, VQAv2, and our SA tasks jointly. 
The number of training examples were 8.6 k, 5.7 k, 393 k, 130 k, 444 k, and 50 k, respectively. The first three tasks are from the GLUE benchmark.

\vspace{0.25em}\noindent\textbf{Compared models.} Table~\ref{table:models} shows the summary. \\
{\bf BERT$_{\rm small}$/+V}: To estimate the upper bound of performance, we fine-tuned BERT$_{\rm small}$ to each task independently, except SA, with task-specific heads. \\
{\bf BERT$_{\rm small}$-s2s+V}:
For comparison with text generation models, we prepared an encoder-decoder model whose encoder and decoder weights were initialized based on the weights of BERT$_{\rm small}$.
We trained that model on all the tasks except for SA jointly.
The input sequences were generated such that they provided the most complete information required to solve a task, for example, task description, question, and class labels, using templates (in Appendix~\ref{app:seq2seq_format}).
The models with the suffix +V use an image input for VQA.
We obtained the grid features using ResNet18 in a manner similar to BUI-BERTs.
We inserted the features into the head of the input embeddings.
Appendix~\ref{app:image_input} provides further details. 

\vspace{0.25em}\noindent\textbf{Setup.}
We trained all models in 10 epochs.
We tracked the validation loss at the end of each epoch to select the best model.
The other conditions for BUI-BERTs were the same as those for the PTA training.
We optimized the hyper-parameters for the compared models (see Appendix~\ref{app:hyperparameter}).

\vspace{0.25em}\noindent\textbf{Results.} Table~\ref{table:results_main} summarizes the results. 
\textcolor{black}{For the non-action or single-page tasks, we estimate the upper bound on the pre-trained BERTs$_{\rm small}$ performance by taking the higher of BERT$_{\rm small}$~(single-task setting) and BERT$_{\rm small}$-s2s+V~(multi-task setting) performance.
As we can see, BUI-BERT$_{\rm small}$ achieves about 76\%-90\% of that estimation except for CoLA in our task framework.
This suggests that pre-trained language models can learn to use their linguistic knowledge in an action sequence, although not perfectly.}
Note that the small CoLA scores in the multi-task setting can be due to the relatively small training data.
\textcolor{black}{
We observed no trend that apply to all tasks in the relationship between model performance and size (BUI-BERT$_{\rm small/meduim}$).
The medium model performs better on the single-page tasks, which is consistent with the size of the language models.
However, their performance is reversed on the multi-page tasks.
Further investigation is required to see if performance scales with model size in our tasks. 
}

\subsection{Analysis} \label{sec:analysis}

\begin{table*}
\small
\centering
\scalebox{1.0}{
\begin{tabular}{lc|ccccc|ccccc}
\hline
{\bf ~} &
{\bf ~} &
{\bf CoLA} &
{\bf STS-B} &
{\bf MNLI-m} &
{\bf VQAv2} & 
{\bf SQuADv2} &
\multicolumn{5}{c}{\bf Search and Answer~(SA)} \\
~ & ~ & ~ & ~ & ~ & ~ & ~ & 
-QID &
-Q &
-H &
-A &
-all \\
\hdashline

~ & \hspace{-5mm} \#steps & 2 & 2 & 2.0 & 5.8 & 9.2 & 23.0 & 30.8 & 13.1 & 19.4 & 24.0  \\
~ & \hspace{-5mm} \#cases & 1043 & 1500 & 9815 & 214354 & 11873 & 1623 & 1634 & 182 & 1561 & 5000 \\

~ & \hspace{-5mm} metric & M & P & macro f1 & acc. & EM & 
acc. & acc. & acc. & acc. & acc.  \\
\hline

\hspace{-2mm} BUI-BERT$_{\rm sml.}$ & \hspace{-5mm}
\begin{tabular}[c]{@{}c@{}} \#sub \\ score \end{tabular} &
\begin{tabular}[c]{@{}c@{}} 1042 \\ -1.0 \end{tabular} &
\begin{tabular}[c]{@{}c@{}} 1500 \\ {\bf 71.6} \end{tabular} &
\begin{tabular}[c]{@{}c@{}} 9815 \\ {\bf 70.4} \end{tabular} &
\begin{tabular}[c]{@{}c@{}} 213236 \\ {\bf 48.0} \end{tabular} &
\begin{tabular}[c]{@{}c@{}} 9563 \\ {\bf 43.6} \end{tabular} &
\begin{tabular}[c]{@{}c@{}} 1592 \\ {\bf 88.8} \end{tabular} &
\begin{tabular}[c]{@{}c@{}} 1615 \\ {\bf 96.0} \end{tabular} &
\begin{tabular}[c]{@{}c@{}} 182 \\ {\bf 100} \end{tabular} &
\begin{tabular}[c]{@{}c@{}} 1514 \\ {\bf 37.2} \end{tabular} &
\begin{tabular}[c]{@{}c@{}} 4903 \\ {\bf 75.5} \end{tabular} \\

~~~~~~w/o PTA & \hspace{-5mm}
\begin{tabular}[c]{@{}c@{}} \#sub \\ score \end{tabular} &
\begin{tabular}[c]{@{}c@{}} 1040 \\ -1.8 \end{tabular} &
\begin{tabular}[c]{@{}c@{}} 1500 \\ 9.3 \end{tabular} &
\begin{tabular}[c]{@{}c@{}} 9813 \\ 68.1 \end{tabular} &
\begin{tabular}[c]{@{}c@{}} 213122 \\ 45.9 \end{tabular} &
\begin{tabular}[c]{@{}c@{}} 9011 \\ 38.9 \end{tabular} &
\begin{tabular}[c]{@{}c@{}} 1559 \\ 60.0 \end{tabular} &
\begin{tabular}[c]{@{}c@{}} 1612 \\ 94.8 \end{tabular} &
\begin{tabular}[c]{@{}c@{}} 182 \\ {\bf 100} \end{tabular} &
\begin{tabular}[c]{@{}c@{}} 1507 \\ 36.1 \end{tabular} &
\begin{tabular}[c]{@{}c@{}} 4860 \\ 65.4 \end{tabular} \\

~~~~~~w/o mem & \hspace{-5mm}
\begin{tabular}[c]{@{}c@{}} \#sub \\ score \end{tabular} &
\begin{tabular}[c]{@{}c@{}} 1043 \\ {\bf 0.0} \end{tabular} &
\begin{tabular}[c]{@{}c@{}} 1500 \\ 51.1 \end{tabular} &
\begin{tabular}[c]{@{}c@{}} 9815 \\ 69.0 \end{tabular} &
\begin{tabular}[c]{@{}c@{}} 212334 \\ 46.9 \end{tabular} &
\begin{tabular}[c]{@{}c@{}} 7669 \\ 27.9 \end{tabular} &
\begin{tabular}[c]{@{}c@{}} 747 \\ 25.3 \end{tabular} &
\begin{tabular}[c]{@{}c@{}} 943 \\ 23.7 \end{tabular} &
\begin{tabular}[c]{@{}c@{}} 182 \\ {\bf 100} \end{tabular} &
\begin{tabular}[c]{@{}c@{}} 1122 \\ 9.7 \end{tabular} &
\begin{tabular}[c]{@{}c@{}} 2994 \\ 22.7 \end{tabular} \\

\hline

\hspace{-2mm} BUI-BERT$_{\rm med.}$ & \hspace{-5mm}
\begin{tabular}[c]{@{}c@{}} \#sub \\ score \end{tabular} &
\begin{tabular}[c]{@{}c@{}} 1040 \\ -1.8 \end{tabular} &
\begin{tabular}[c]{@{}c@{}} 1500 \\ 78.3 \end{tabular} &
\begin{tabular}[c]{@{}c@{}} 9815 \\ 75.6 \end{tabular} &
\begin{tabular}[c]{@{}c@{}} 213599 \\ 48.8 \end{tabular} &
\begin{tabular}[c]{@{}c@{}} 10082 \\ 49.4 \end{tabular} &
\begin{tabular}[c]{@{}c@{}} 1591 \\ 93.7 \end{tabular} &
\begin{tabular}[c]{@{}c@{}} 1614 \\ 96.0 \end{tabular} &
\begin{tabular}[c]{@{}c@{}} 182 \\ 100 \end{tabular} &
\begin{tabular}[c]{@{}c@{}} 1506 \\ 13.8 \end{tabular} &
\begin{tabular}[c]{@{}c@{}} 4893 \\ 69.7 \end{tabular} \\

\hline
\end{tabular}
}
\vspace{-2mm}
\caption{\label{table:results_ablation}
\textcolor{black}{
Ablation study with the validation splits.
\#steps : the averaged numbers of steps in the gold sequences.
\#cases : the number of cases evaluated.
\#sub : the number of cases where the model made a submission.
We counted the cases with no submission as failure cases.
M and P represent Matthews' and Pearson's correlation, respectively.
}}
\end{table*}

\begin{table}
\vspace{-4mm}
\small
\centering
\scalebox{1.0}{
\begin{tabular}{l|ccc}
\hline
(\#cor / \#sub) & {\bf WNLI} & {\bf MRPC} & {\bf SST-2} \\
\hdashline
~~~~~~~~~~~~~~~~~~~\#cases & {71} & {408} & {872} \\
\hline
BERT$_{\rm small}$-s2s+V & 8 / 14 & 0 / 0  & 1 / 2\\
BUI-BERT$_{\rm small}$ & 17 / 33 & 181 / 227 & 120 / 271 \\
BUI-BERT$_{\rm medium}$ & 4 / 9 & 9 /29  & 29 / 53 \\
\hline
\end{tabular}
}
\vspace{-2mm}
\caption{\label{table:results_zeroshot}
\textcolor{black}{
Unseen task evaluation on the validation splits.
\#sub (\#cor) : the number of cases where the model was successful in submitting an answer (a correct answer).
\#cases : the number of cases evaluated.}}
\end{table}

\noindent\textbf{Ablation study.}
We added two models to validate PTA and the memory mechanism.
For BUI-BERT$_{\rm small}$ w/o PTA, we initialized its weight with BERT$_{\rm small}$ and directly trained it on the multi-task training.
For BUI-BERT$_{\rm small}$ w/o mem, we omitted the memory sequence. 
This model was re-initialized and trained on PTA.
Table~\ref{table:results_ablation} shows the results.
Ablated models were lower than BUI-BERT$_{\rm small}$ on almost all the tasks.
This shows that PTA and the memory mechanism is effective.
Especially, the significant degradation of BUI-BERT$_{\rm small}$ w/o mem on the SA tasks suggests that memory plays an important role.

\vspace{0.25em}\noindent\textbf{SA tasks.}
\textcolor{black}{
We present the result of SA tasks in detail in Table~\ref{table:results_ablation}.
As we can see, all models were successful in submitting an answer in most cases except for BUI-BERT$_{\rm small}$ w/o mem. 
However, the SA-A task scores are low.
This low scores reflect the difference of the tasks: while the SA-QID, -Q, and -H tasks can be solved by copying displayed results, the SA-A task requires inferring the answer for a question.
Bridging this gap is a challenge for the future.
Note that the scores on SA-QID, -Q and -H also reflect the difficulty: 
SA-H is the easiest because the answer (the hit count) is always displayed at the first line; 
SA-Q is relatively easy because the hit count is always;
SA-QID is more difficult because the hit count is more than one, and models are sometimes required to find the answer by scrolling.
There is still room for improvement regarding the SA-Q and -QID tasks. 
}

\vspace{0.25em}\noindent\textbf{Unseen tasks.}
\textcolor{black}{
To see how the models understood the instruction, we evaluated them on unseen tasks.
}
We used three GLUE tasks: WNLI~\cite{levesque2012winograd}, MRPC~\cite{dolan2005automatically}, and SST-2~\cite{socher2013recursive}.
Those were two-choice tasks, and their similarity to the learned tasks was differed.
WNLI and MNLI were textual entailment tasks.
MRPC and STS-B were equivalence and similarity tasks.
SST-2 is a sentiment prediction, which was new to the models.
We report the number of submissions~(\#sub) and correct submissions~(\#cor) in Table~\ref{table:results_zeroshot}.
\textcolor{black}{
We observed that BUI-BERT$_{\rm small\/medium}$ failed to submit an answer in many cases although the submission rate of those models is above a chance level estimated from the percentage of the area occupied by the buttons on the screen.
Based on the percentage of correct answers in the submissions, it does not seem that the models refrain from submitting their answers when they are not confident.
This low submission rate suggests that the models likely memorize the entire procedures without aligning the actions in a procedure to the words in an instruction.
}

\section{Discussion} \label{sec:discussion}
\textcolor{black}{Expanding the world that NLP can handle is one of our long-term goals~\cite{bisk2020experience}.
Models that manipulate GUIs provide a unified basis for studies on dynamic interactions, e.g., dynamic grounding~\cite{chandu-etal-2021-grounding}, because such models can adapt to diverse software through GUIs.
We converted existing benchmark tasks into GUI tasks in this paper.
Such GUI tasks would help evaluate how well the performance of a language model is transferred to a GUI model.
This evaluation viewpoint complements MiniWoB~\cite{shi2017world} and MiniWoB++~\cite{liu2018reinforcement}.
}

\textcolor{black}{Although our experiments showed encouraging results that language models can learn GUI manipulation while retaining linguistic knowledge to some extent, our study has the following limitations.
\begin{itemize}[leftmargin=*]
\setlength{\itemsep}{-3pt}
\item \vspace{-4pt} The high computational cost precludes detailed experiments, including validation of scale by using larger LMs. 
We need to seek light-weight models, such as efficient transformers~\cite{tay2020efficient} or Perceiver~\cite{pmlr-v139-jaegle21a}.
\item Performance on unseen tasks is low. Although we expect that data augmentation~(i.e., modifying the page layout) can improve the performance, reinforcement learning~\cite{sutton2018reinforcement} is likely to be essential for exploration.
\item Our models use the emulated OCR, which relies on a browser. 
To apply our models to GUIs other than the browser GUI, changing that function remains future work.
While using an actual OCR is a straightforward solution, its processing time is not negligible for our purpose. 
Therefore, we might create LMs that directly read from pixels.
\end{itemize}}

\section{Conclusion} \label{sec:conclusion}

In this work, we demonstrated that BERT can be applied to a task framework that requires multiple actions to use a browser UI.
In multi-task training, our BERT extension with a memory mechanism learned to solve six tasks according to the UI, including hyperlinks, provided by the task pages.
Simultaneously, we observed the low ability to solve unseen tasks.
Because the computational cost of current models limits in-depth experiments, we plan to create light-weighted models with existing efficient methods as the next step.

\section*{Acknowledgments}
This work was supported by JSPS KAKENHI Grant Number 21H03502 and JST SPRING Grant Number JPMJSP2104.

\bibliographystyle{acl_natbib}
\bibliography{custom}

\begin{thebibliography}{59}
\expandafter\ifx\csname natexlab\endcsname\relax\def\natexlab#1{#1}\fi

\bibitem[{Anderson et~al.(2018)Anderson, Wu, Teney, Bruce, Johnson,
  S{\"u}nderhauf, Reid, Gould, and Van Den~Hengel}]{anderson2018vision}
Peter Anderson, Qi~Wu, Damien Teney, Jake Bruce, Mark Johnson, Niko
  S{\"u}nderhauf, Ian Reid, Stephen Gould, and Anton Van Den~Hengel. 2018.
\newblock \href
  {https://openaccess.thecvf.com/content_cvpr_2018/html/Anderson_Vision-and-Language_Navigation_Interpreting_CVPR_2018_paper.html}
  {{Vision-and-Language Navigation: Interpreting Visually-Grounded Navigation
  Instructions in Real Environments}}.
\newblock In \emph{Proceedings of the IEEE Conference on Computer Vision and
  Pattern Recognition}, pages 3674--3683.

\bibitem[{Antol et~al.(2015)Antol, Agrawal, Lu, Mitchell, Batra, Zitnick, and
  Parikh}]{antol2015vqa}
Stanislaw Antol, Aishwarya Agrawal, Jiasen Lu, Margaret Mitchell, Dhruv Batra,
  C.~Lawrence Zitnick, and Devi Parikh. 2015.
\newblock \href
  {https://openaccess.thecvf.com/content_iccv_2015/html/Antol_VQA_Visual_Question_ICCV_2015_paper.html}
  {{VQA: Visual Question Answering}}.
\newblock In \emph{Proceedings of the IEEE International Conference on Computer
  Vision}, pages 2425--2433.

\bibitem[{Bai et~al.(2021)Bai, Zang, Xu, Sunkara, Rastogi, Chen, and Agüera~y
  Arcas}]{baiuibert}
Chongyang Bai, Xiaoxue Zang, Ying Xu, Srinivas Sunkara, Abhinav Rastogi,
  Jindong Chen, and Blaise Agüera~y Arcas. 2021.
\newblock \href {https://doi.org/10.24963/ijcai.2021/235} {{UIBert: Learning
  Generic Multimodal Representations for UI Understanding}}.
\newblock In \emph{Proceedings of the Thirtieth International Joint Conference
  on Artificial Intelligence}, pages 1705--1712.

\bibitem[{Bisk et~al.(2020)Bisk, Holtzman, Thomason, Andreas, Bengio, Chai,
  Lapata, Lazaridou, May, Nisnevich, Pinto, and Turian}]{bisk2020experience}
Yonatan Bisk, Ari Holtzman, Jesse Thomason, Jacob Andreas, Yoshua Bengio, Joyce
  Chai, Mirella Lapata, Angeliki Lazaridou, Jonathan May, Aleksandr Nisnevich,
  Nicolas Pinto, and Joseph Turian. 2020.
\newblock \href {https://doi.org/10.18653/v1/2020.emnlp-main.703} {{Experience
  Grounds Language}}.
\newblock In \emph{Proceedings of the 2020 Conference on Empirical Methods in
  Natural Language Processing}, pages 8718--8735.

\bibitem[{Brown et~al.(2020)Brown, Mann, Ryder, Subbiah, Kaplan, Dhariwal,
  Neelakantan, Shyam, Sastry, Askell, Agarwal, Herbert-Voss, Krueger, Henighan,
  Child, Ramesh, Ziegler, Wu, Winter, Hesse, Chen, Sigler, Litwin, Gray, Chess,
  Clark, Berner, McCandlish, Radford, Sutskever, and
  Amodei}]{brown2020language}
Tom~B. Brown, Benjamin Mann, Nick Ryder, Melanie Subbiah, Jared Kaplan,
  Prafulla Dhariwal, Arvind Neelakantan, Pranav Shyam, Girish Sastry, Amanda
  Askell, Sandhini Agarwal, Ariel Herbert-Voss, Gretchen Krueger, T.~J.
  Henighan, Rewon Child, Aditya Ramesh, Daniel~M. Ziegler, Jeff Wu, Clemens
  Winter, Christopher Hesse, Mark Chen, Eric Sigler, Mateusz Litwin, Scott
  Gray, Benjamin Chess, Jack Clark, Christopher Berner, Sam McCandlish, Alec
  Radford, Ilya Sutskever, and Dario Amodei. 2020.
\newblock \href {https://arxiv.org/abs/2005.14165} {{Language Models are
  Few-Shot Learners}}.
\newblock \emph{arXiv preprint arXiv:2005.14165}.

\bibitem[{Caruana(1997)}]{caruana1997multitask}
Rich Caruana. 1997.
\newblock \href {https://link.springer.com/article/10.1023/A:1007379606734}
  {{Multitask Learning}}.
\newblock \emph{Machine Learning}, 28:pages 41--75.

\bibitem[{Cer et~al.(2017)Cer, Diab, Agirre, Lopez-Gazpio, and
  Specia}]{cer2017semeval}
Daniel Cer, Mona Diab, Eneko Agirre, I{\~n}igo Lopez-Gazpio, and Lucia Specia.
  2017.
\newblock \href {https://doi.org/10.18653/v1/S17-2001} {{SemEval-2017 Task 1:
  Semantic Textual Similarity Multilingual and Crosslingual Focused
  Evaluation}}.
\newblock In \emph{Proceedings of the 11th International Workshop on Semantic
  Evaluation}, pages 1--14.

\bibitem[{Chandu et~al.(2021)Chandu, Bisk, and
  Black}]{chandu-etal-2021-grounding}
Khyathi~Raghavi Chandu, Yonatan Bisk, and Alan~W Black. 2021.
\newblock \href {https://doi.org/10.18653/v1/2021.findings-acl.375} {{Grounding
  `Grounding' in NLP}}.
\newblock In \emph{Findings of the Association for Computational Linguistics:
  ACL-IJCNLP 2021}, pages 4283--4305.

\bibitem[{Chen et~al.(2021)Chen, Zhao, Chen, Ji, Zhang, Luo, Xiong, and
  Yu}]{chen2021websrc}
Xingyu Chen, Zihan Zhao, Lu~Chen, JiaBao Ji, Danyang Zhang, Ao~Luo, Yuxuan
  Xiong, and Kai Yu. 2021.
\newblock \href {https://aclanthology.org/2021.emnlp-main.343} {{WebSRC: A
  Dataset for Web-Based Structural Reading Comprehension}}.
\newblock In \emph{Proceedings of the 2021 Conference on Empirical Methods in
  Natural Language Processing}, pages 4173--4185.

\bibitem[{Chen et~al.(2020)Chen, Li, Yu, Kholy, Ahmed, Gan, Cheng, and
  Liu}]{chen2020uniter}
Yen-Chun Chen, Linjie Li, Licheng Yu, Ahmed~El Kholy, Faisal Ahmed, Zhe Gan,
  Yu~Cheng, and Jingjing Liu. 2020.
\newblock \href {https://link.springer.com/chapter/10.1007/978-3-030-58577-8_7}
  {{UNITER: Learning UNiversal Image-TExt Representations}}.
\newblock In \emph{The 2020 European Conference on Computer Vision}, pages
  104--120.

\bibitem[{Cho et~al.(2021)Cho, Lei, Tan, and Bansal}]{cho2021unifying}
Jaemin Cho, Jie Lei, Hao Tan, and Mohit Bansal. 2021.
\newblock \href {https://proceedings.mlr.press/v139/cho21a.html} {{Unifying
  Vision-and-Language Tasks via Text Generation}}.
\newblock In \emph{Proceedings of the 38th International Conference on Machine
  Learning}, volume 139, pages 1931--1942.

\bibitem[{Cui et~al.(2021)Cui, Xu, Lv, and Wei}]{cui2021document}
Lei Cui, Yiheng Xu, Tengchao Lv, and Furu Wei. 2021.
\newblock \href {https://arxiv.org/abs/2111.08609} {{Document AI: Benchmarks,
  Models and Applications}}.
\newblock \emph{arXiv preprint arXiv:2111.08609}.

\bibitem[{Das et~al.(2018)Das, Datta, Gkioxari, Lee, Parikh, and
  Batra}]{das2018embodied}
Abhishek Das, Samyak Datta, Georgia Gkioxari, Stefan Lee, Devi Parikh, and
  Dhruv Batra. 2018.
\newblock \href
  {https://openaccess.thecvf.com/content_cvpr_2018/html/Das_Embodied_Question_Answering_CVPR_2018_paper.html}
  {{Embodied Question Answering}}.
\newblock In \emph{Proceedings of the IEEE Conference on Computer Vision and
  Pattern Recognition}, pages 1--10.

\bibitem[{Devlin et~al.(2019)Devlin, Chang, Lee, and
  Toutanova}]{devlin2019bert}
Jacob Devlin, Ming-Wei Chang, Kenton Lee, and Kristina Toutanova. 2019.
\newblock \href {https://doi.org/10.18653/v1/N19-1423} {{BERT: Pre-training of
  Deep Bidirectional Transformers for Language Understanding}}.
\newblock In \emph{Proceedings of the 2019 Conference of the North {A}merican
  Chapter of the Association for Computational Linguistics: Human Language
  Technologies, Volume 1}, pages 4171--4186.

\bibitem[{Dolan and Brockett(2005)}]{dolan2005automatically}
William~B. Dolan and Chris Brockett. 2005.
\newblock \href {https://www.aclweb.org/anthology/I05-5002} {{Automatically
  Constructing a Corpus of Sentential Paraphrases}}.
\newblock In \emph{Proceedings of the Third International Workshop on
  Paraphrasing}.

\bibitem[{Gur et~al.(2021)Gur, Jaques, Miao, Choi, Tiwari, Lee, and
  Faust}]{gur2021environment}
Izzeddin Gur, Natasha Jaques, Yingjie Miao, Jongwook Choi, Manoj Tiwari,
  Honglak Lee, and Aleksandra Faust. 2021.
\newblock \href {https://openreview.net/forum?id=CeByDMy0YTL} {Environment
  generation for zero-shot compositional reinforcement learning}.
\newblock In \emph{Advances in Neural Information Processing Systems}.

\bibitem[{Gur et~al.(2018)Gur, Rueckert, Faust, and
  Hakkani-Tur}]{gur2018learning}
Izzeddin Gur, Ulrich Rueckert, Aleksandra Faust, and Dilek Hakkani-Tur. 2018.
\newblock \href {https://openreview.net/forum?id=BJemQ209FQ} {Learning to
  navigate the web}.
\newblock In \emph{International Conference on Learning Representations}.

\bibitem[{He et~al.(2016)He, Zhang, Ren, and Sun}]{he2016deep}
Kaiming He, Xiangyu Zhang, Shaoqing Ren, and Jian Sun. 2016.
\newblock \href
  {https://openaccess.thecvf.com/content_cvpr_2016/html/He_Deep_Residual_Learning_CVPR_2016_paper.html}
  {{Deep Residual Learning for Image Recognition}}.
\newblock In \emph{Proceedings of the IEEE conference on computer vision and
  pattern recognition}, pages 770--778.

\bibitem[{He et~al.(2021)He, Sunkara, Zang, Xu, Liu, Wichers, Schubiner, Lee,
  and Chen}]{he2021actionbert}
Zecheng He, Srinivas Sunkara, Xiaoxue Zang, Ying Xu, Lijuan Liu, Nevan Wichers,
  Gabriel Schubiner, Ruby Lee, and Jindong Chen. 2021.
\newblock \href {https://ojs.aaai.org/index.php/AAAI/article/view/16741}
  {{ActionBert: Leveraging User Actions for Semantic Understanding of User
  Interfaces}}.
\newblock In \emph{Proceedings of the AAAI Conference on Artificial
  Intelligence}, volume~35, pages 5931--5938.

\bibitem[{Hong et~al.(2021)Hong, Wu, Qi, Rodriguez-Opazo, and
  Gould}]{hong2021vln}
Yicong Hong, Qi~Wu, Yuankai Qi, Cristian Rodriguez-Opazo, and Stephen Gould.
  2021.
\newblock \href
  {https://openaccess.thecvf.com/content/CVPR2021/html/Hong_VLN_BERT_A_Recurrent_Vision-and-Language_BERT_for_Navigation_CVPR_2021_paper.html}
  {{VLN BERT: A Recurrent Vision-and-Language BERT for Navigation}}.
\newblock In \emph{Proceedings of the IEEE/CVF Conference on Computer Vision
  and Pattern Recognition}, pages 1643--1653.

\bibitem[{Hu and Singh(2021)}]{hu2021unit}
Ronghang Hu and Amanpreet Singh. 2021.
\newblock \href
  {https://openaccess.thecvf.com/content/ICCV2021/html/Hu_UniT_Multimodal_Multitask_Learning_With_a_Unified_Transformer_ICCV_2021_paper.html}
  {{UniT: Multimodal Multitask Learning With a Unified Transformer}}.
\newblock In \emph{Proceedings of the IEEE/CVF International Conference on
  Computer Vision (ICCV)}, pages 1439--1449.

\bibitem[{Huang et~al.(2020)Huang, Zeng, Liu, Fu, and Fu}]{huang2020pixel}
Zhicheng Huang, Zhaoyang Zeng, Bei Liu, Dongmei Fu, and Jianlong Fu. 2020.
\newblock \href {https://arxiv.org/abs/2004.00849} {{Pixel-BERT: Aligning Image
  Pixels with Text by Deep Multi-Modal Transformers}}.
\newblock \emph{arXiv preprint arXiv:2004.00849}.

\bibitem[{Humphreys et~al.(2022)Humphreys, Raposo, Pohlen, Thornton, Chhaparia,
  Muldal, Abramson, Georgiev, Goldin, Santoro, and
  Lillicrap}]{humphreys2022data}
Peter~C. Humphreys, David Raposo, Tobias Pohlen, Gregory Thornton, Rachita
  Chhaparia, Alistair Muldal, Josh Abramson, Petko Georgiev, Alex Goldin, Adam
  Santoro, and Timothy~P. Lillicrap. 2022.
\newblock \href {https://arxiv.org/abs/2202.08137} {A data-driven approach for
  learning to control computers}.
\newblock \emph{arXiv preprint arXiv:2202.08137}.

\bibitem[{Jaegle et~al.(2021)Jaegle, Gimeno, Brock, Vinyals, Zisserman, and
  Carreira}]{pmlr-v139-jaegle21a}
Andrew Jaegle, Felix Gimeno, Andy Brock, Oriol Vinyals, Andrew Zisserman, and
  Joao Carreira. 2021.
\newblock \href {https://proceedings.mlr.press/v139/jaegle21a.html} {Perceiver:
  General perception with iterative attention}.
\newblock In \emph{Proceedings of the 38th International Conference on Machine
  Learning}, volume 139 of \emph{Proceedings of Machine Learning Research},
  pages 4651--4664.

\bibitem[{Jia et~al.(2019)Jia, Kiros, and Ba}]{jia2018domqnet}
Sheng Jia, Jamie~Ryan Kiros, and Jimmy Ba. 2019.
\newblock \href {https://openreview.net/forum?id=HJgd1nAqFX} {{DOM}-q-{NET}:
  Grounded {RL} on structured language}.
\newblock In \emph{International Conference on Learning Representations}.

\bibitem[{Kazemzadeh et~al.(2014)Kazemzadeh, Ordonez, Matten, and
  Berg}]{kazemzadeh2014referitgame}
Sahar Kazemzadeh, Vicente Ordonez, Mark Matten, and Tamara Berg. 2014.
\newblock \href {https://aclanthology.org/D14-1086/} {Referitgame: Referring to
  objects in photographs of natural scenes}.
\newblock In \emph{Proceedings of the 2014 conference on empirical methods in
  natural language processing (EMNLP)}, pages 787--798.

\bibitem[{Kingma and Ba(2014)}]{kingma2014adam}
Diederik~P Kingma and Jimmy Ba. 2014.
\newblock \href {https://arxiv.org/abs/1412.6980} {{Adam: A Method for
  Stochastic Optimization}}.
\newblock \emph{arXiv preprint arXiv:1412.6980}.

\bibitem[{Levesque et~al.(2012)Levesque, Davis, and
  Morgenstern}]{levesque2012winograd}
Hector Levesque, Ernest Davis, and Leora Morgenstern. 2012.
\newblock \href {https://dl.acm.org/doi/10.5555/3031843.3031909} {{The Winograd
  Schema Challenge}}.
\newblock In \emph{Thirteenth International Conference on the Principles of
  Knowledge Representation and Reasoning}.

\bibitem[{Li et~al.(2021{\natexlab{a}})Li, Xu, Cui, and Wei}]{li2021markuplm}
Junlong Li, Yiheng Xu, Lei Cui, and Furu Wei. 2021{\natexlab{a}}.
\newblock \href {https://arxiv.org/abs/2110.08518} {{MarkupLM: Pre-training of
  Text and Markup Language for Visually-rich Document Understanding}}.
\newblock \emph{arXiv preprint arXiv:2110.08518}.

\bibitem[{Li et~al.(2019{\natexlab{a}})Li, Yatskar, Yin, Hsieh, and
  Chang}]{li2019visualbert}
Liunian~Harold Li, Mark Yatskar, Da~Yin, Cho-Jui Hsieh, and Kai-Wei Chang.
  2019{\natexlab{a}}.
\newblock \href {https://arxiv.org/abs/1908.03557} {{VisualBERT: A simple and
  performant baseline for vision and language}}.
\newblock \emph{arXiv preprint arXiv:1908.03557}.

\bibitem[{Li et~al.(2019{\natexlab{b}})Li, Li, Xia, Bisk, Celikyilmaz, Gao,
  Smith, and Choi}]{li2019robust}
Xiujun Li, Chunyuan Li, Qiaolin Xia, Yonatan Bisk, Asli Celikyilmaz, Jianfeng
  Gao, Noah~A Smith, and Yejin Choi. 2019{\natexlab{b}}.
\newblock \href {https://aclanthology.org/D19-1159/} {{Robust Navigation with
  Language Pretraining and Stochastic Sampling}}.
\newblock In \emph{Proceedings of the 2019 Conference on Empirical Methods in
  Natural Language Processing and the 9th International Joint Conference on
  Natural Language Processing (EMNLP-IJCNLP)}, pages 1494--1499.

\bibitem[{Li et~al.(2021{\natexlab{b}})Li, Li, Zhou, Dehghani, and
  Gritsenko}]{li2021vut}
Yang Li, Gang Li, Xin Zhou, Mostafa Dehghani, and Alexey Gritsenko.
  2021{\natexlab{b}}.
\newblock \href {https://arxiv.org/abs/2112.05692} {{VUT: Versatile UI
  Transformer for Multi-Modal Multi-Task User Interface Modeling}}.
\newblock \emph{arXiv preprint arXiv:2112.05692}.

\bibitem[{Liu et~al.(2018)Liu, Guu, Pasupat, Shi, and
  Liang}]{liu2018reinforcement}
Evan~Zheran Liu, Kelvin Guu, Panupong Pasupat, Tianlin Shi, and Percy Liang.
  2018.
\newblock \href {https://openreview.net/forum?id=ryTp3f-0-} {Reinforcement
  learning on web interfaces using workflow-guided exploration}.
\newblock In \emph{International Conference on Learning Representations}.

\bibitem[{Lu et~al.(2019)Lu, Batra, Parikh, and Lee}]{NEURIPS2019_c74d97b0}
Jiasen Lu, Dhruv Batra, Devi Parikh, and Stefan Lee. 2019.
\newblock \href
  {https://papers.nips.cc/paper/2019/hash/c74d97b01eae257e44aa9d5bade97baf-Abstract.html}
  {{ViLBERT: Pretraining Task-Agnostic Visiolinguistic Representations for
  Vision-and-Language Tasks}}.
\newblock In \emph{Advances in Neural Information Processing Systems},
  volume~32.

\bibitem[{Lu et~al.(2020)Lu, Goswami, Rohrbach, Parikh, and Lee}]{lu202012}
Jiasen Lu, Vedanuj Goswami, Marcus Rohrbach, Devi Parikh, and Stefan Lee. 2020.
\newblock \href
  {https://openaccess.thecvf.com/content_CVPR_2020/html/Lu_12-in-1_Multi-Task_Vision_and_Language_Representation_Learning_CVPR_2020_paper.html}
  {{12-in-1: Multi-Task Vision and Language Representation Learning}}.
\newblock In \emph{Proceedings of the IEEE/CVF Conference on Computer Vision
  and Pattern Recognition}, pages 10437--10446.

\bibitem[{Majumdar et~al.(2020)Majumdar, Shrivastava, Lee, Anderson, Parikh,
  and Batra}]{majumdar2020improving}
Arjun Majumdar, Ayush Shrivastava, Stefan Lee, Peter Anderson, Devi Parikh, and
  Dhruv Batra. 2020.
\newblock \href
  {https://www.ecva.net/papers/eccv_2020/papers_ECCV/html/5672_ECCV_2020_paper.php}
  {{Improving Vision-and-Language Navigation with Image-Text Pairs from the
  Web}}.
\newblock In \emph{European Conference on Computer Vision}, pages 259--274.

\bibitem[{Merity et~al.(2016)Merity, Xiong, Bradbury, and
  Socher}]{merity2016pointer}
Stephen Merity, Caiming Xiong, James Bradbury, and Richard Socher. 2016.
\newblock \href {https://arxiv.org/abs/1609.07843} {Pointer sentinel mixture
  models}.
\newblock \emph{arXiv preprint arXiv:1609.07843}.

\bibitem[{Qi et~al.(2021)Qi, Pan, Hong, Yang, van~den Hengel, and
  Wu}]{Qi_2021_ICCV}
Yuankai Qi, Zizheng Pan, Yicong Hong, Ming-Hsuan Yang, Anton van~den Hengel,
  and Qi~Wu. 2021.
\newblock \href
  {https://openaccess.thecvf.com/content/ICCV2021/html/Qi_The_Road_To_Know-Where_An_Object-and-Room_Informed_Sequential_BERT_for_ICCV_2021_paper.html}
  {{The Road To Know-Where: An Object-and-Room Informed Sequential BERT for
  Indoor Vision-Language Navigation}}.
\newblock In \emph{Proceedings of the IEEE/CVF International Conference on
  Computer Vision (ICCV)}, pages 1655--1664.

\bibitem[{Radford et~al.(2018)Radford, Narasimhan, Salimans, and
  Sutskever}]{radford2018improving}
Alec Radford, Karthik Narasimhan, Tim Salimans, and Ilya Sutskever. 2018.
\newblock \href {https://openai.com/blog/language-unsupervised/} {Improving
  language understanding by generative pre-training}.
\newblock \emph{OpenAI Blog}.

\bibitem[{Radford et~al.(2019)Radford, Wu, Child, Luan, Amodei, and
  Sutskever}]{radford2019language}
Alec Radford, Jeff Wu, Rewon Child, David Luan, Dario Amodei, and Ilya
  Sutskever. 2019.
\newblock \href
  {https://d4mucfpksywv.cloudfront.net/better-language-models/language_models_are_unsupervised_multitask_learners.pdf}
  {{Language Models are Unsupervised Multitask Learners}}.
\newblock \emph{OpenAI blog}, 1(8).

\bibitem[{Raffel et~al.(2020)Raffel, Shazeer, Roberts, Lee, Narang, Matena,
  Zhou, Li, and Liu}]{raffel2020exploring}
Colin Raffel, Noam Shazeer, Adam Roberts, Katherine Lee, Sharan Narang, Michael
  Matena, Yanqi Zhou, Wei Li, and Peter~J Liu. 2020.
\newblock \href {https://jmlr.org/papers/v21/20-074.html} {{Exploring the
  Limits of Transfer Learning with a Unified Text-to-Text Transformer}}.
\newblock \emph{Journal of Machine Learning Research}, 21(140):pages 1--67.

\bibitem[{Rajpurkar et~al.(2018)Rajpurkar, Jia, and Liang}]{rajpurkar2018know}
Pranav Rajpurkar, Robin Jia, and Percy Liang. 2018.
\newblock \href {https://aclanthology.org/P18-2124/} {{Know What You Don’t
  Know: Unanswerable Questions for SQuAD}}.
\newblock In \emph{Proceedings of the 56th Annual Meeting of the Association
  for Computational Linguistics Volume 2}, pages 784--789.

\bibitem[{Rajpurkar et~al.(2016)Rajpurkar, Zhang, Lopyrev, and
  Liang}]{rajpurkar2016squad}
Pranav Rajpurkar, Jian Zhang, Konstantin Lopyrev, and Percy Liang. 2016.
\newblock \href {https://doi.org/10.18653/v1/D16-1264} {{{SQuAD: 100,000+
  Questions for Machine Comprehension of Text}}}.
\newblock In \emph{Proceedings of the 2016 Conference on Empirical Methods in
  Natural Language Processing}, pages 2383--2392.

\bibitem[{Rothe et~al.(2020)Rothe, Narayan, and Severyn}]{rothe2020leveraging}
Sascha Rothe, Shashi Narayan, and Aliaksei Severyn. 2020.
\newblock \href {https://aclanthology.org/2020.tacl-1.18/} {{Leveraging
  Pre-Trained Checkpoints for Sequence Generation Tasks}}.
\newblock \emph{Transactions of the Association for Computational Linguistics},
  8:pages 264--280.

\bibitem[{Ruder(2017)}]{ruder2017overview}
Sebastian Ruder. 2017.
\newblock \href {https://arxiv.org/abs/1706.05098} {{An Overview of Multi-Task
  Learning in Deep Neural Networks}}.
\newblock \emph{arXiv preprint arXiv:1706.05098}.

\bibitem[{Shi et~al.(2017)Shi, Karpathy, Fan, Hernandez, and
  Liang}]{shi2017world}
Tianlin Shi, Andrej Karpathy, Linxi Fan, Jonathan Hernandez, and Percy Liang.
  2017.
\newblock \href {http://proceedings.mlr.press/v70/shi17a.html} {World of bits:
  An open-domain platform for web-based agents}.
\newblock In \emph{International Conference on Machine Learning}, pages
  3135--3144.

\bibitem[{Shridhar et~al.(2020)Shridhar, Thomason, Gordon, Bisk, Han, Mottaghi,
  Zettlemoyer, and Fox}]{shridhar2020alfred}
Mohit Shridhar, Jesse Thomason, Daniel Gordon, Yonatan Bisk, Winson Han,
  Roozbeh Mottaghi, Luke Zettlemoyer, and Dieter Fox. 2020.
\newblock \href
  {https://openaccess.thecvf.com/content_CVPR_2020/html/Shridhar_ALFRED_A_Benchmark_for_Interpreting_Grounded_Instructions_for_Everyday_Tasks_CVPR_2020_paper.html}
  {{ALFRED: A Benchmark for Interpreting Grounded Instructions for Everyday
  Tasks}}.
\newblock In \emph{Proceedings of the IEEE/CVF conference on computer vision
  and pattern recognition}, pages 10740--10749.

\bibitem[{Socher et~al.(2013)Socher, Perelygin, Wu, Chuang, Manning, Ng, and
  Potts}]{socher2013recursive}
Richard Socher, Alex Perelygin, Jean Wu, Jason Chuang, Christopher~D. Manning,
  Andrew Ng, and Christopher Potts. 2013.
\newblock \href {https://www.aclweb.org/anthology/D13-1170} {{Recursive Deep
  Models for Semantic Compositionality Over a Sentiment Treebank}}.
\newblock In \emph{Proceedings of the 2013 Conference on Empirical Methods in
  Natural Language Processing}, pages 1631--1642.

\bibitem[{Srivastava et~al.(2020)Srivastava, Polozov, Jojic, and
  Meek}]{srivastava-etal-2020-learning}
Shashank Srivastava, Oleksandr Polozov, Nebojsa Jojic, and Christopher Meek.
  2020.
\newblock \href {https://doi.org/10.18653/v1/2020.acl-main.684} {Learning
  web-based procedures by reasoning over explanations and demonstrations in
  context}.
\newblock In \emph{Proceedings of the 58th Annual Meeting of the Association
  for Computational Linguistics}, pages 7652--7662, Online. Association for
  Computational Linguistics.

\bibitem[{Su et~al.(2020)Su, Zhu, Cao, Li, Lu, Wei, and Dai}]{Su2020VL-BERT:}
Weijie Su, Xizhou Zhu, Yue Cao, Bin Li, Lewei Lu, Furu Wei, and Jifeng Dai.
  2020.
\newblock \href {https://openreview.net/forum?id=SygXPaEYvH} {{VL-BERT:
  Pre-training of Generic Visual-Linguistic Representations}}.
\newblock In \emph{International Conference on Learning Representations}.

\bibitem[{Sutton and Barto(2018)}]{sutton2018reinforcement}
Richard~S Sutton and Andrew~G Barto. 2018.
\newblock \emph{Reinforcement learning: An introduction}.
\newblock MIT press.

\bibitem[{Tan and Bansal(2019)}]{tan-bansal-2019-lxmert}
Hao Tan and Mohit Bansal. 2019.
\newblock \href {https://doi.org/10.18653/v1/D19-1514} {{{LXMERT: Learning
  Cross-Modality Encoder Representations from Transformers}}}.
\newblock In \emph{Proceedings of the 2019 Conference on Empirical Methods in
  Natural Language Processing and the 9th International Joint Conference on
  Natural Language Processing}, pages 5100--5111.

\bibitem[{Tanaka et~al.(2021)Tanaka, Nishida, and
  Yoshida}]{tanaka2021visualmrc}
Ryota Tanaka, Kyosuke Nishida, and Sen Yoshida. 2021.
\newblock \href {https://ojs.aaai.org/index.php/AAAI/article/view/17635}
  {{VisualMRC: Machine Reading Comprehension on Document Images}}.
\newblock In \emph{Proceedings of the AAAI Conference on Artificial
  Intelligence}, volume~35, pages 13878--13888.

\bibitem[{Tay et~al.(2020)Tay, Dehghani, Bahri, and Metzler}]{tay2020efficient}
Yi~Tay, Mostafa Dehghani, Dara Bahri, and Donald Metzler. 2020.
\newblock \href {https://arxiv.org/abs/2009.06732} {Efficient transformers: A
  survey}.
\newblock \emph{arXiv preprint arXiv:2009.06732}.

\bibitem[{Vaswani et~al.(2017)Vaswani, Shazeer, Parmar, Uszkoreit, Jones,
  Gomez, Kaiser, and Polosukhin}]{vaswani2017attention}
Ashish Vaswani, Noam Shazeer, Niki Parmar, Jakob Uszkoreit, Llion Jones,
  Aidan~N Gomez, {\L}ukasz Kaiser, and Illia Polosukhin. 2017.
\newblock \href
  {https://papers.nips.cc/paper/2017/hash/3f5ee243547dee91fbd053c1c4a845aa-Abstract.html}
  {{Attention Is All You Need}}.
\newblock In \emph{Advances in neural information processing systems}, pages
  5998--6008.

\bibitem[{Wang et~al.(2019)Wang, Singh, Michael, Hill, Levy, and
  Bowman}]{wang2019glue}
Alex Wang, Amanpreet Singh, Julian Michael, Felix Hill, Omer Levy, and
  Samuel~R. Bowman. 2019.
\newblock \href {https://openreview.net/forum?id=rJ4km2R5t7} {{GLUE: A
  Multi-Task Benchmark and Analysis Platform for Natural Language
  Understanding}}.
\newblock In \emph{International Conference on Learning Representations}.

\bibitem[{Warstadt et~al.(2019)Warstadt, Singh, and
  Bowman}]{warstadt2019neural}
Alex Warstadt, Amanpreet Singh, and Samuel~R. Bowman. 2019.
\newblock \href {https://doi.org/10.1162/tacl_a_00290} {{Neural Network
  Acceptability Judgments}}.
\newblock \emph{Transactions of the Association for Computational Linguistics},
  7:pages 625--641.

\bibitem[{Williams et~al.(2018)Williams, Nangia, and
  Bowman}]{williams2018broad}
Adina Williams, Nikita Nangia, and Samuel Bowman. 2018.
\newblock \href {https://doi.org/10.18653/v1/N18-1101} {{A Broad-Coverage
  Challenge Corpus for Sentence Understanding through Inference}}.
\newblock In \emph{Proceedings of the 2018 Conference of the North {A}merican
  Chapter of the Association for Computational Linguistics: Human Language
  Technologies, Volume 1}, pages 1112--1122.

\bibitem[{Wu et~al.(2021)Wu, Li, Zhang, Chen, Hombaiah, and
  Bendersky}]{wu2021lampret}
Te-Lin Wu, Cheng Li, Mingyang Zhang, Tao Chen, Spurthi~Amba Hombaiah, and
  Michael Bendersky. 2021.
\newblock \href {https://arxiv.org/abs/2104.08405} {{LAMPRET: Layout-Aware
  Multimodal PreTraining for Document Understanding}}.
\newblock \emph{arXiv preprint arXiv:2104.08405}.

\end{thebibliography}

\newpage
\appendix

\section{Environment Detail}

\paragraph{Browser.} We used the following environment to render and execute task pages:
\begin{itemize}
\setlength{\itemsep}{-2pt}
\item \vspace{-4pt} OS: Ubuntu 18.04, 20.04
\item Browser: Firefox version 87.0
\item Browser driver: geckodriver 0.29.0
\item Selenium: version 3.141.0
\item Default font (main text): Dejavu Serif, 16px
\item Default font (text, button): 13px
\end{itemize}

\paragraph{Packages and libraries.} We used Python 3.6 and PyTorch (1.10) to implement our BUI models.
For sequence to sequence models, we used the Transformers library (4.12).
To evaluate SQuAD and VQA, we used the public scripts\footnote{SQuAD : \url{https://rajpurkar.github.io/SQuAD-explorer/} and VQA : \url{https://github.com/GT-Vision-Lab/VQA}.} .
 
\paragraph{Training.} We used a NVIDIA V100 GPU with 32 GB VRAM or a NVIDIA RTX 3090 GPU with 24 GM VRAM in each training run.

\subsection{Required Time}

\begin{table*}
\small
\centering
\begin{tabular}{lll}
\hline
process & time & remarks \\
\hline
record gold seqs for PTA & 6h & 62k examples. \\
record gold seqs for the others & $\sim$4d & $\sim$1.3M examples. \\
\hline
Train small / medium on PTA in 50 ep. & $\sim$2d / $\sim$4d & 60k examples. with a GPU. \\
Train small / medium on the multi-task training in 10 ep. & $\sim$6d / $\sim$12d & $\sim$1.0M examples. with a GPU. \\
Predict with a model on val. split of PTA & 20min & 2k examples. with a GPU. \\
Predict with a model on val. split of the others & $\sim$2d & $\sim$230k examples. with a GPU. \\
\hline
\end{tabular}
\caption{\label{table:required_time}
Required time.
Since we used several servers with the different configurations, those values are approximations.
Gold sequences are reusable if the screensize, tokenization and actions of the models are identical.
We saved the screenshots and actions of a single example in a single json file, and read it from disks each time we used it.
We used float32 for the training.
We also tried float16 with automated mix precision.
Although it reduced the training time by about 30\% (we doubled the batchsize using the reduced memory space), it sometimes caused NaNs and stop the training.
Therefore, we did not use it this time.
}
\end{table*}

We report the required time for our experiments in the Table~\ref{table:required_time}.

\section{Additional Details on Compared Models}

\subsection{Input Templates} \label{app:seq2seq_format}

We used templates to make the text inputs for the seq2seq models.
Table~\ref{table:seq2seq_input} shows the templates and the rule to fill the sentences in the template for each dataset.
We made the templates so that they provided task\_type, instruction, content names and content values.
In addition, image embeddings are added to the head of sequence for VQA.
We made TASK\_TYPE and INSTRUCTION with reference to the contents of the data set.

\begin{sidewaystable*}

\begin{tabular}{ll}
\hline
\#contents & template \\
\hline
1 &
[TASK\_TYPE]~:~[INSTRUCTION]~[VALUE\_1] \\
2 &
[TASK\_TYPE]~:~[INSTRUCTION]~[KEY\_1]~=~[VALUE\_1]~[KEY\_2]~=~[VALUE\_2] \\
\hline
\end{tabular}{}
\vspace{15pt}~~ \\
\begin{tabular}{llp{9cm}llll}
\hline
 & TASK\_TYPE & INSTRUCTION & KEY\_1 & VALUE\_1 & KEY\_2 & VALUE\_2 \\
\hline
VQA & visual question answering & See the picture and answer the following question. & question & (question) &  &  \\
SQuAD & question and answering & Read the next paragraph and answer the following question. answer an empty string when you think the question is unanswerable. & Paragraph & (paragraph) & Question & (question) \\
CoLA & single choice classification & If the following sentence is acceptable as an English sentence, answer acceptable; if not, answer unacceptable. & sentence & (sentence) &  &  \\
SST-2 & single choice classification & Predict the emotion of the sentence (positive / negative). & sentence & (sentence) &  &  \\
STS-B & single choice classification & Rate how similar the following two sentences are on a scale from 0 to 5 (0 being the least similar and  being the most similar 5). & sentence1 & (sentence1) & sentence2 & (sentence2) \\
MRPC & single choice classification & Answer whether the following pairs of sentences are semantically equivalent. If they are equivalent, answer equivalent; if not, answer not equivalent. & sentence1 & (\#1 String) & sentence2 & (\#2 String) \\
MNLI & single choice classification & For the following premise and hypothesis statements, answer entailment if the premise entails the hypothesis, contradiction if it contradicts the hypothesis, or neutral if neither. & Premise & (sentence1) & Hypothesis & (sentence2) \\
WNLI & single choice classification & Read the following two sentences and answer their relationship: enntailment or not entailment. & sentence1 & (sentence1) & sentence2 & (sentence2) \\
\hline
\end{tabular}
\caption{\label{table:seq2seq_input}
(top) the templates for text input, and (bottom) the rules to fill the sentences to the templates for seq2seq models.
We used data from the datasets for the value fields.
}
\end{sidewaystable*}

\subsection{Image Input} \label{app:image_input}
First, we resize an given image 320px$\times$~224px. 
If the aspect ratio does not match, we center the image and fill in the missing pixels with black pixel. 
Second, we input the image to the frozen pre-trained ResNet18 model to obtain the last feature map (C4; 10x7).
We flat the feature map in one dimension and input each feature to one fully-connected linear layer, which is trainable, to  align the dimension with the hidden dimension of the LM. 
Finally, we concatenate those features and language embeddings before adding positional and segment type embeddings.

\subsection{Hyper-parameters} \label{app:hyperparameter}

We used the ADAM optimizer without scheduling of learning rate (LR), and enabled Automatic Mixed Precision (AMP).
Every training was 10 epoch.
In a preliminary experiment, we observed that optimization of pre-trained ResNet18 had little impact on the VQA performance, so we only used the frozen setting above.

\paragraph{BERT$_{\rm small}$ / +V.} 
We fixed the max token length for the GLUE tasks and VQA 300, and for SQuAd 512.
We tried six hyper-parameter combination: the mini-batch size from \{64, 128, 256\}, the LR from \{1e-4, 5e-5\}.
We adopted the hyper-parameter set whose smallest validation loss was the smallest.

\paragraph{BERT$_{\rm small}$-s2s+V.}
We fixed the base mini-batch size 32, the max token length for text-only tasks 512 and for text-and-image tasks 432 (+70 image embeddings).
We tried six hyper-parameter combination: gradient accumulation from \{1, 4\} and LR from \{1e-4, 5e-5, 1e-5\}.
We adopted the hyper-parameter set whose smallest validation loss was the smallest.
The best hyper-parameters were (4, 5e-5) for BERT$_{\rm small}$-s2s+V

\subsection{Classification with Seq2Seq models}
For classification tasks, we considered the model failed to submit an answer when the generated text did not exactly match any class labels specified in the instruction.

\section{Tasks in the BUI setup}

\subsection{Instructions for Answer Forms} \label{app_instructions}

\begin{figure*}
\centering
\includegraphics[width=\linewidth]{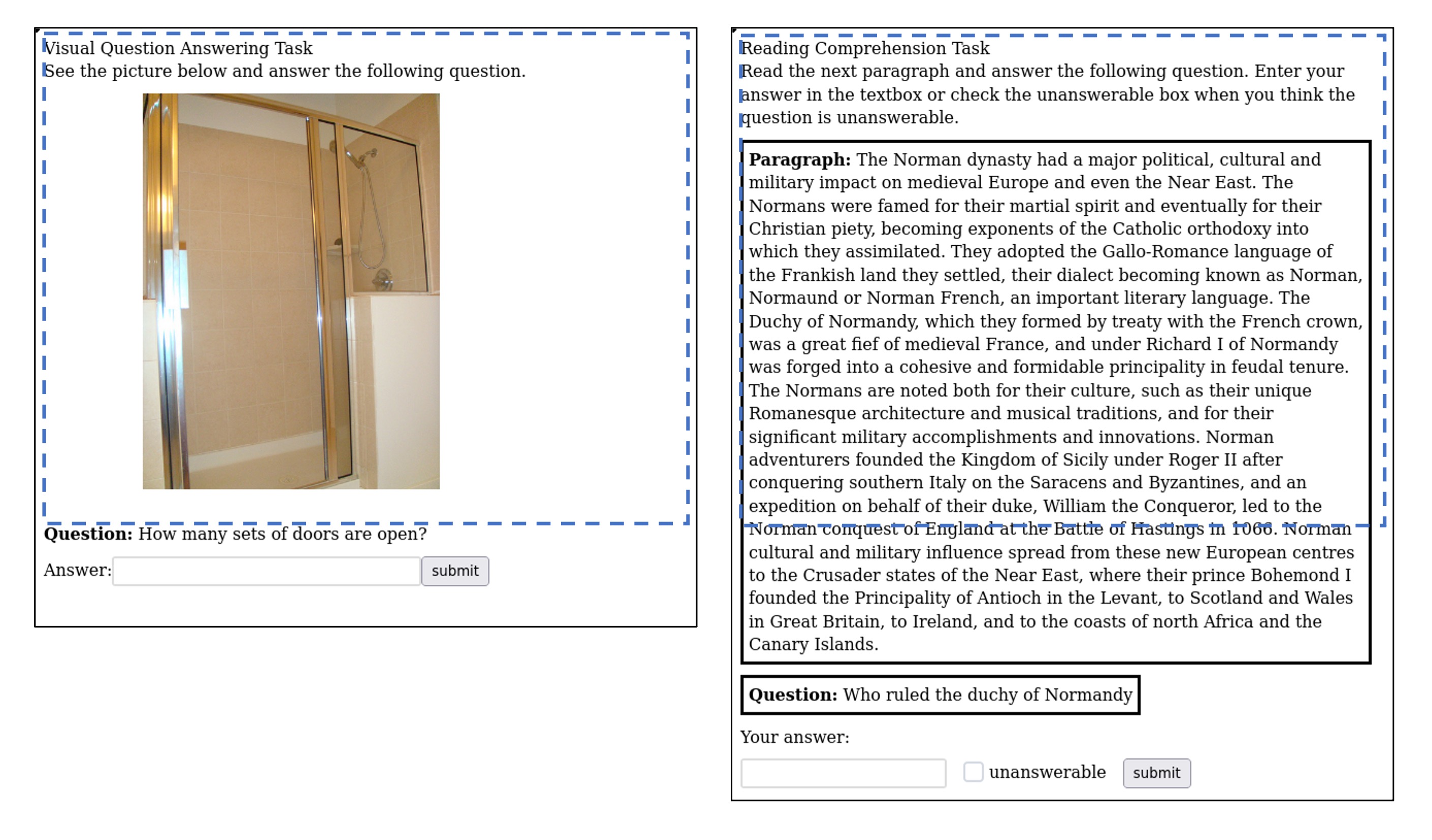}
\vspace{-2mm}
\caption{\label{figure:task_qa}
Screen examples form the BUI version of VQAv2~(left) and that of SQuADv2~(right).
Those are screenshot that the BUI models receive.
The blue dash rectangles show the initial visible area for the models.
The instructions and answer forms are common for the all examples.
}
\end{figure*}

\begin{figure*}
\centering
\includegraphics[width=\linewidth]{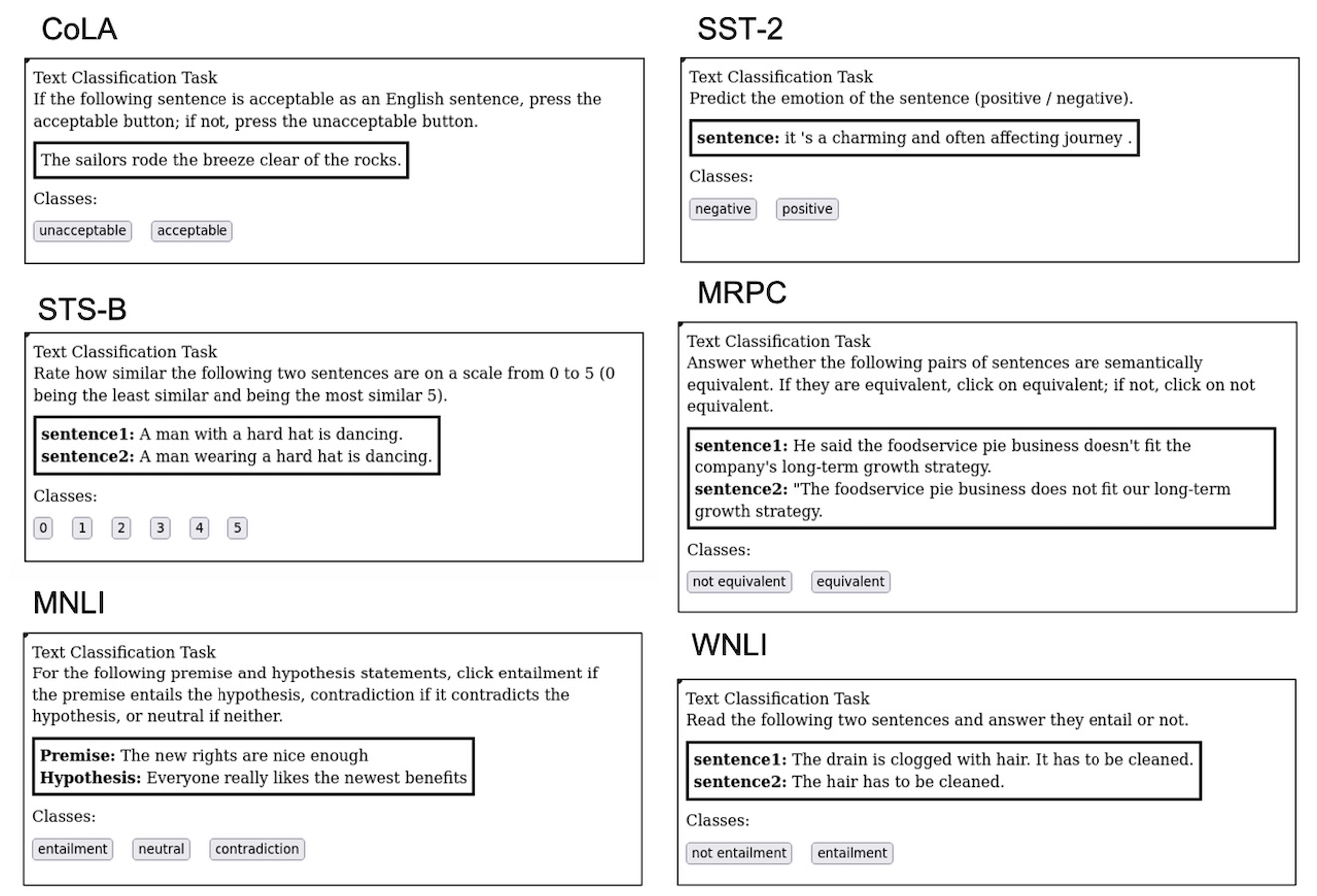}
\vspace{-2mm}
\caption{\label{figure:task_doc_class}
Screen examples from the BUI version of the GLUE benchmark.
The bottom margins are omitted.
While the contents in the bold solid boxes change depend on the examples, the instruction and the label buttons are common.
Note that tasks we did not used (QNLI, QQP, and RTE) are not presented.
}
\end{figure*}

\begin{figure*}
\centering
\includegraphics[width=\linewidth]{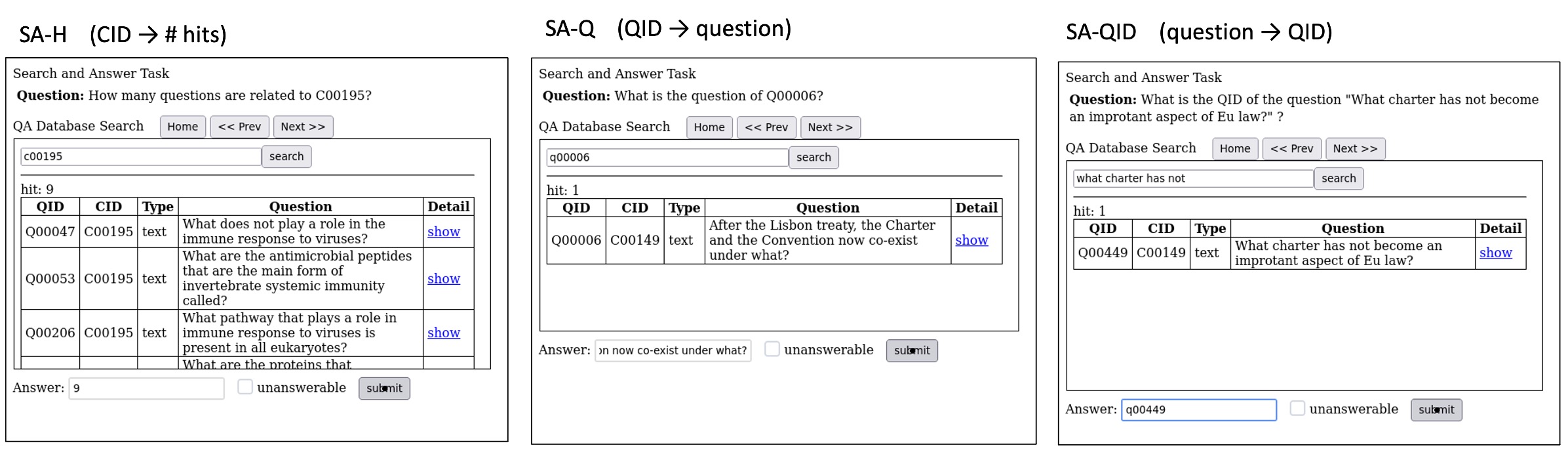}
\vspace{-2mm}
\caption{\label{figure:task_sa_1}
Screen examples of SA-QID, -Q and -H.
The screenshots show the last step of tasks. 
These tasks are expected to be solved by (1) extracting a key phrase from a given instruction, (2) querying the key phrase, (3) finding an answer segment, and (4) entering the segment. 
}
\end{figure*}

\begin{figure*}
\centering
\includegraphics[width=\linewidth]{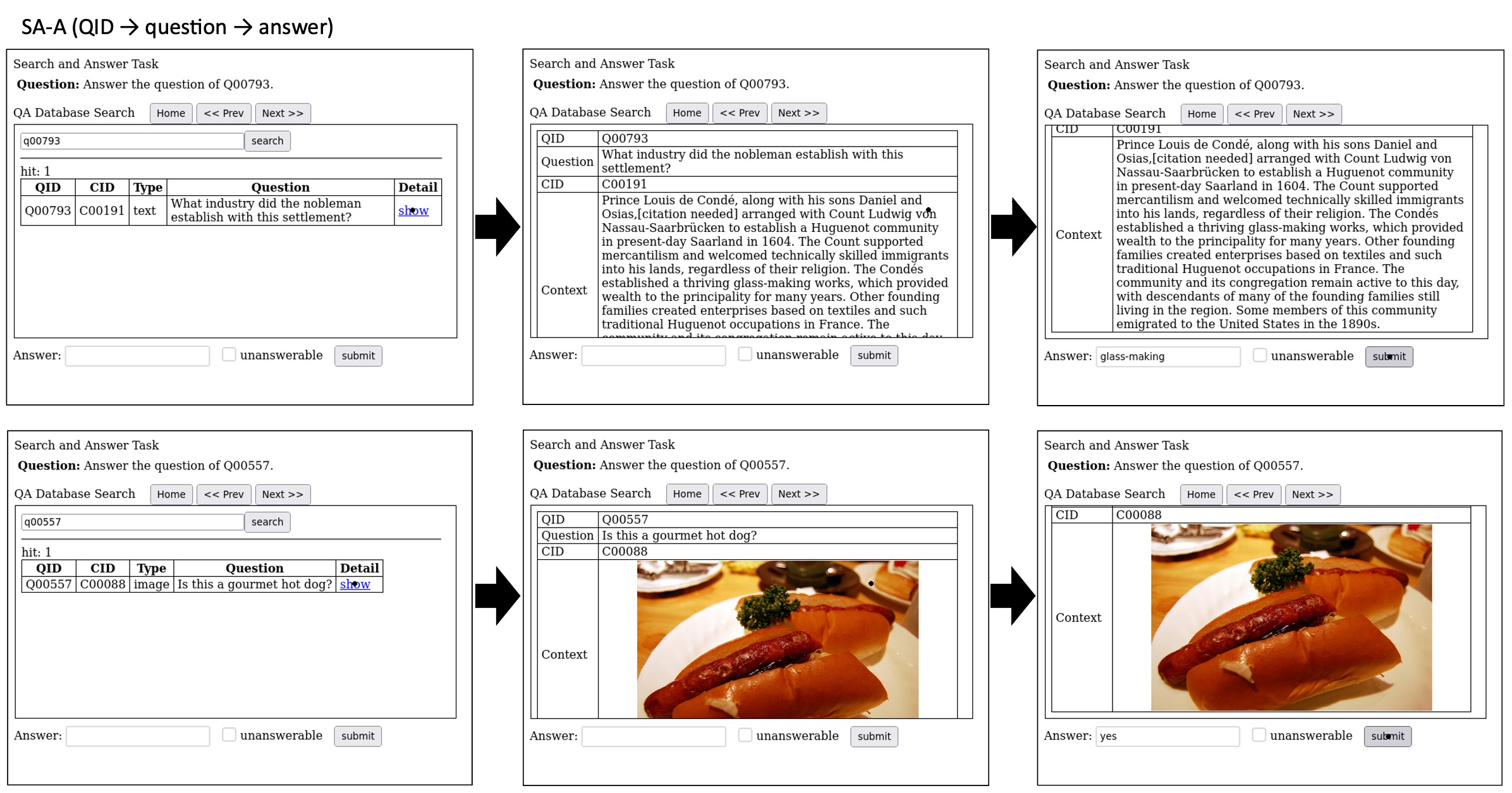}
\vspace{-2mm}
\caption{\label{figure:task_sa_2}
Screen examples of SA-A.
These tasks are expected to be solved by (1) extracting a key phrase from a given instruction, (2) querying the key phrase, (3) showing the detail, (4) reading the question, (5) finding the answer in the context, and (6) entering the answer.
}
\end{figure*}

Here, we shows the instructions and answer forms as images.
Figure~\ref{figure:task_qa} shows the SQuAD and VQA pages.
Figure~\ref{figure:task_doc_class} shows the task pages for the GLUE tasks.
Figure~\ref{figure:task_sa_1} and Figure~\ref{figure:task_sa_2} show the task pages for the SA tasks.
Instructions are basically the same as the counter parts for the seq2seq models shown in Table~\ref{table:seq2seq_input} except for that word choices are changed so that they fit to the screen.

\subsection{Templates for Pre-Training for Actions} \label{app_pta_templates}

\paragraph{Vocabulary.}
We made a vocabulary from the training split of the Wikitext103~\cite{merity2016pointer} corpus.
We kept the words that consist of only alphabets and numbers.
We lower-cased the words.

Sets of words in the instructions are expanded to make the variation.
We sampled one uniformly from the instructions for a task instance.

\subsubsection{Cursor}
\paragraph{Instructions:}
\begin{itemize}
\setlength{\itemsep}{-2pt}
\item \vspace{-4pt} Move the cursor in the box.
\item Point to the box with the cursor.
\end{itemize}
The coordinates of the box was sampled from a window uniformly.

\subsubsection{Button}
\paragraph{Instructions:}
\begin{itemize}
\setlength{\itemsep}{-2pt}
\item \vspace{-4pt} \{Click, Push, Press, Choose, Select\} the button labelled WORD.
\item \{Click, Push, Press, Choose, Select\} the WORD button.
\end{itemize}
WORD was sampled from the vocabulary.

\subsubsection{Area}
\paragraph{Instructions:}
\begin{itemize}
\setlength{\itemsep}{-2pt}
\item \vspace{-4pt} Scroll down until the buttons appear and click the button labelled WORD.
\item Scroll down until the buttons appear and click the WORD button.
\end{itemize}
WORD was sampled from the vocabulary.

\subsubsection{Text}
\paragraph{Instructions:}
\begin{itemize}
\setlength{\itemsep}{-2pt}
\item \vspace{-4pt} \{Type, Enter, Input\} the string to the left of it in each text box. Click the submit button at last.
\end{itemize}
Each string was made by jointing two words, sampled from the vocabulary, with a space.

\subsection{Detail of Search and Answer Tasks}\label{app:sa}
For Search and Answer Tasks, we sampled 100 contexts (paragraphs or images) from each of SQuAD and VQA to create a database.
The database contains $\sim$2k questions because each context has approximately 10 questions.
We chose this database size to make it difficult to enter the whole data into the model.
We assigned unique labels to each context and question in the database, CID, and QID, and created four tasks.
A database yields 200 SA-H tasks and $\sim$2k SA-QID, -Q, -A tasks.
Finally, we sampled 500 tasks from those generated tasks.

In total, we created 100 databases (50000 tasks) for the training split, and 10 databases (5000 tasks) for the validation split.
The contexts do not overlap between databases.

The search UI uses partial matching on the entries

\subsection{Distribution of the Gold Sequence Length}

\begin{figure}
\centering
\includegraphics[width=\linewidth]{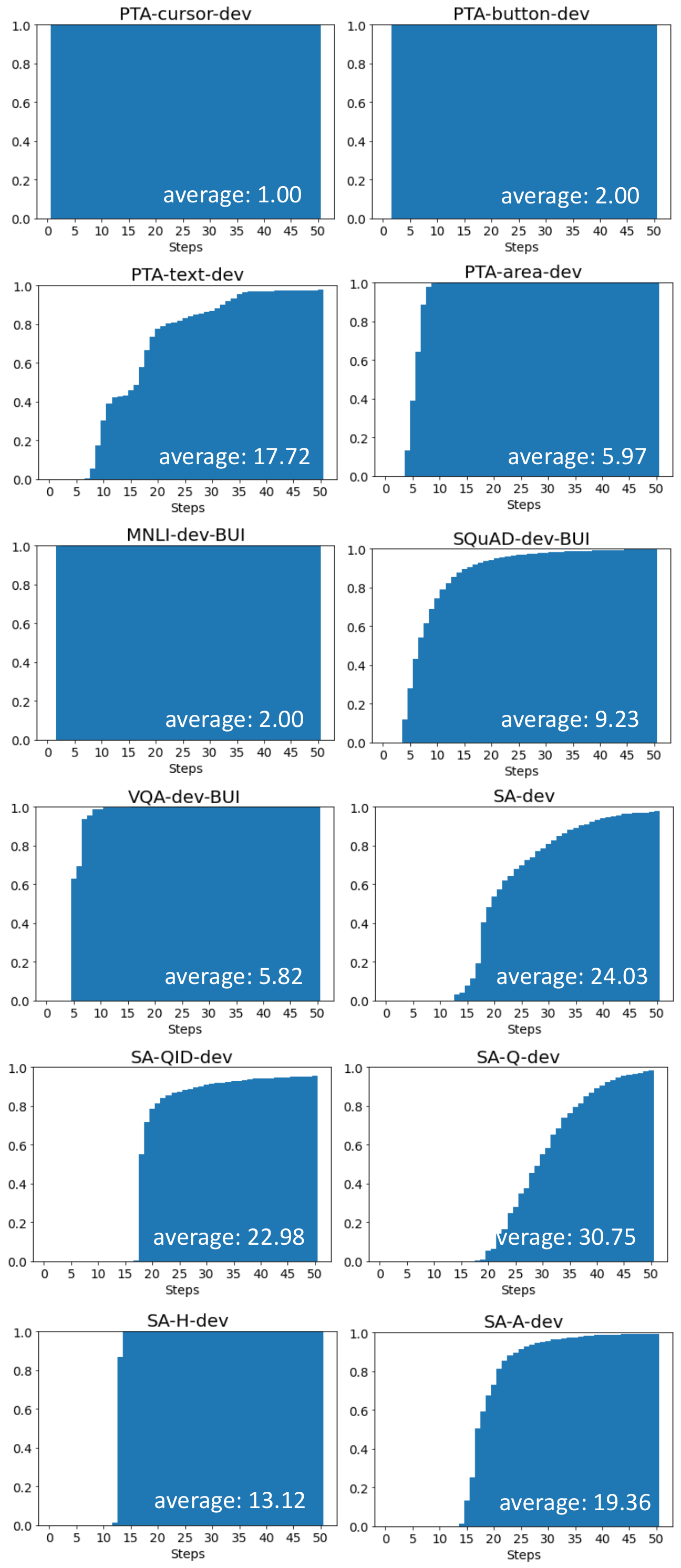}
\vspace{-2mm}
\caption{\label{figure:action_dist}
Distributions of the length of gold action sequences on the dev splits.
We show cumulative values.
Since the number of actions in the document classification task is basically two, we showed MNLI as a typical example.
}
\end{figure}

Figure~\ref{figure:action_dist} shows the distributions of the length of gold action sequences.
Almost all of the examples fall within the upper limit of 50 steps that we set during our training.
Tasks that require entering answers into text boxes tend to have a longer number of steps.

\section{Additional Details on BUI-BERTs}

\subsection{OCR Emulation} \label{app_detection}

\begin{figure*}
\centering
\includegraphics[width=0.95\linewidth]{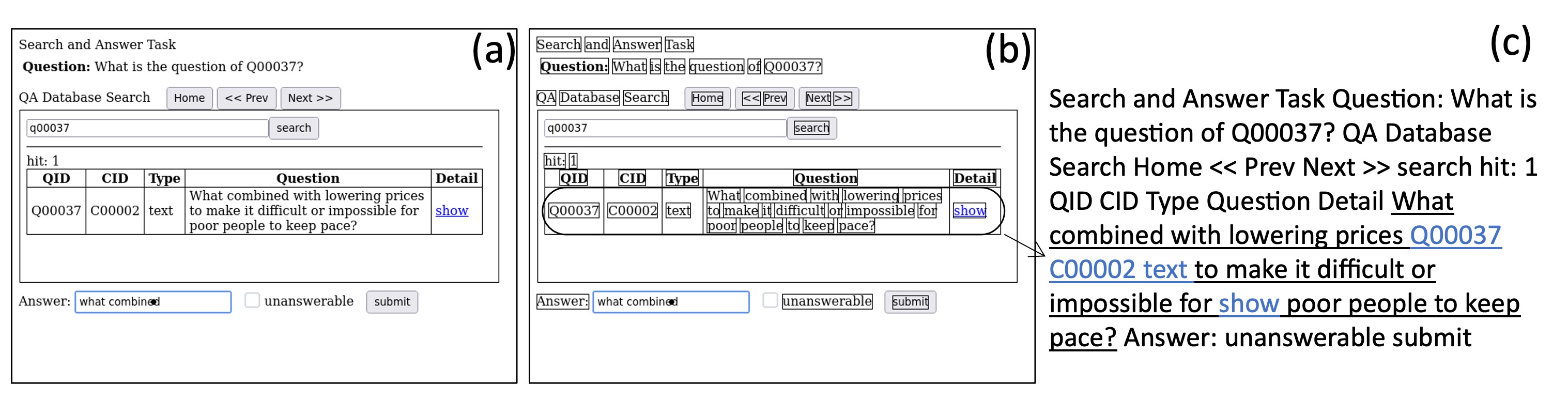}
\vspace{-2mm}
\caption{\label{figure:ocr_emulation}
Example of our OCR emulation.
(a) Example screen. 
(b) Detected words. detected words are surrounded by solid boxes. 
(c) Obtained text sequence. Parts with the broken order are underlined.
}
\end{figure*}

We used OCR emulation, where we surrounds each word in HTML sources using span tags, instead of real OCR in this work.
Figure~\ref{figure:ocr_emulation} shows an example.
The Emulation do not capture the text in text boxes owing to technical reason.
Words are sorted in a top faster and left faster manner.
Sorting preserves natural orders basically, but it sometimes breaks the order as shown in the figure.

\subsection{Mini-Batching Strategy}
\label{mb_setup}

\begin{figure}
\centering
\includegraphics[width=0.95\linewidth]{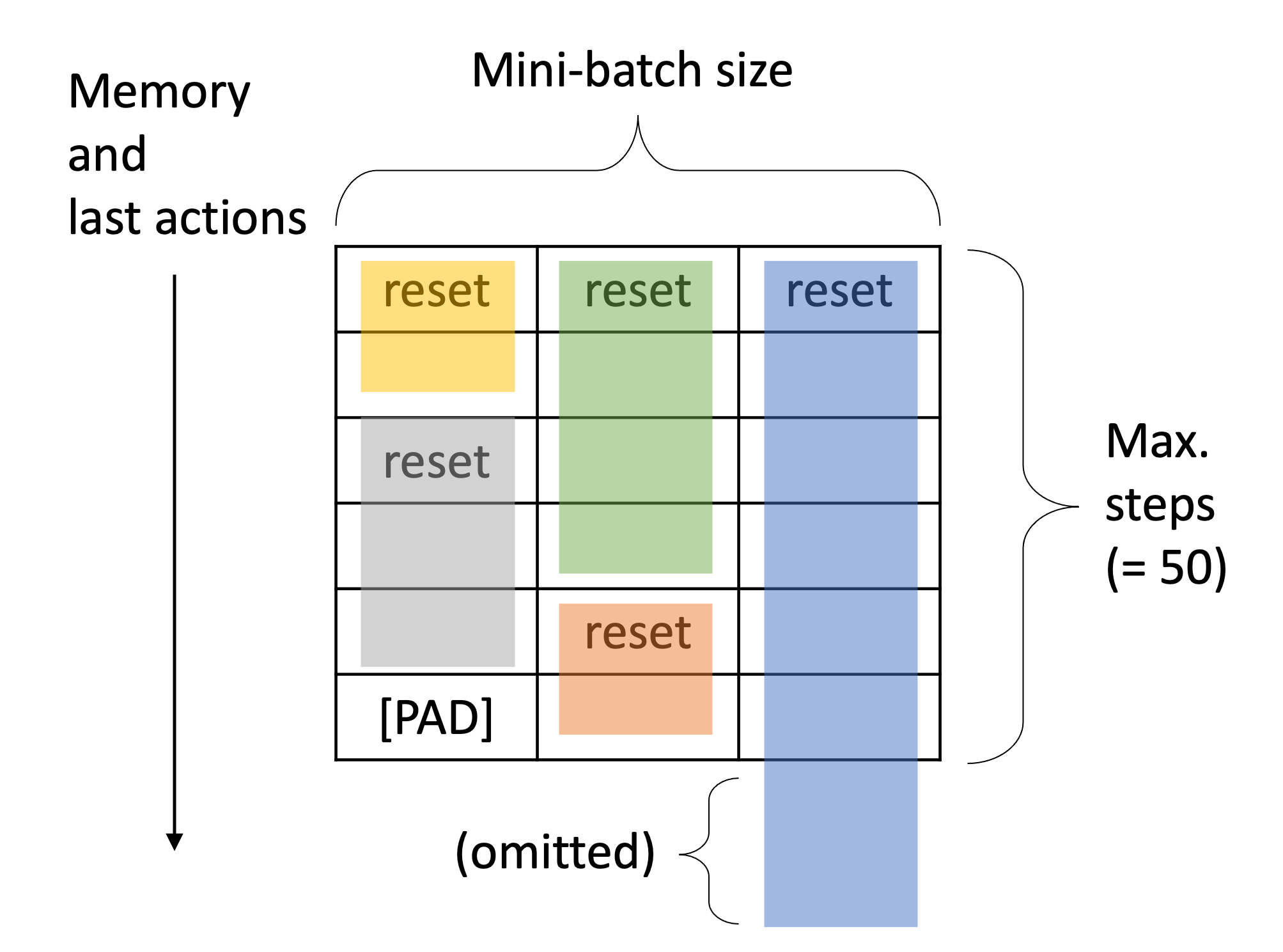}
\vspace{-2mm}
\caption{\label{figure:minibatch}
Mini-batching for multi-step training.
}
\end{figure}

Figure~\ref{figure:minibatch} illustrates mini-batching we used for training.
We packed multiple trajectories in a line of mini-batches to increase the filling rate.
We input memory and last actions recurrently for a trajectory and reset them at each head of trajectories.

\subsection{Learning Curves of BUI-BERTs}

\begin{figure}
\centering
\includegraphics[width=\linewidth]{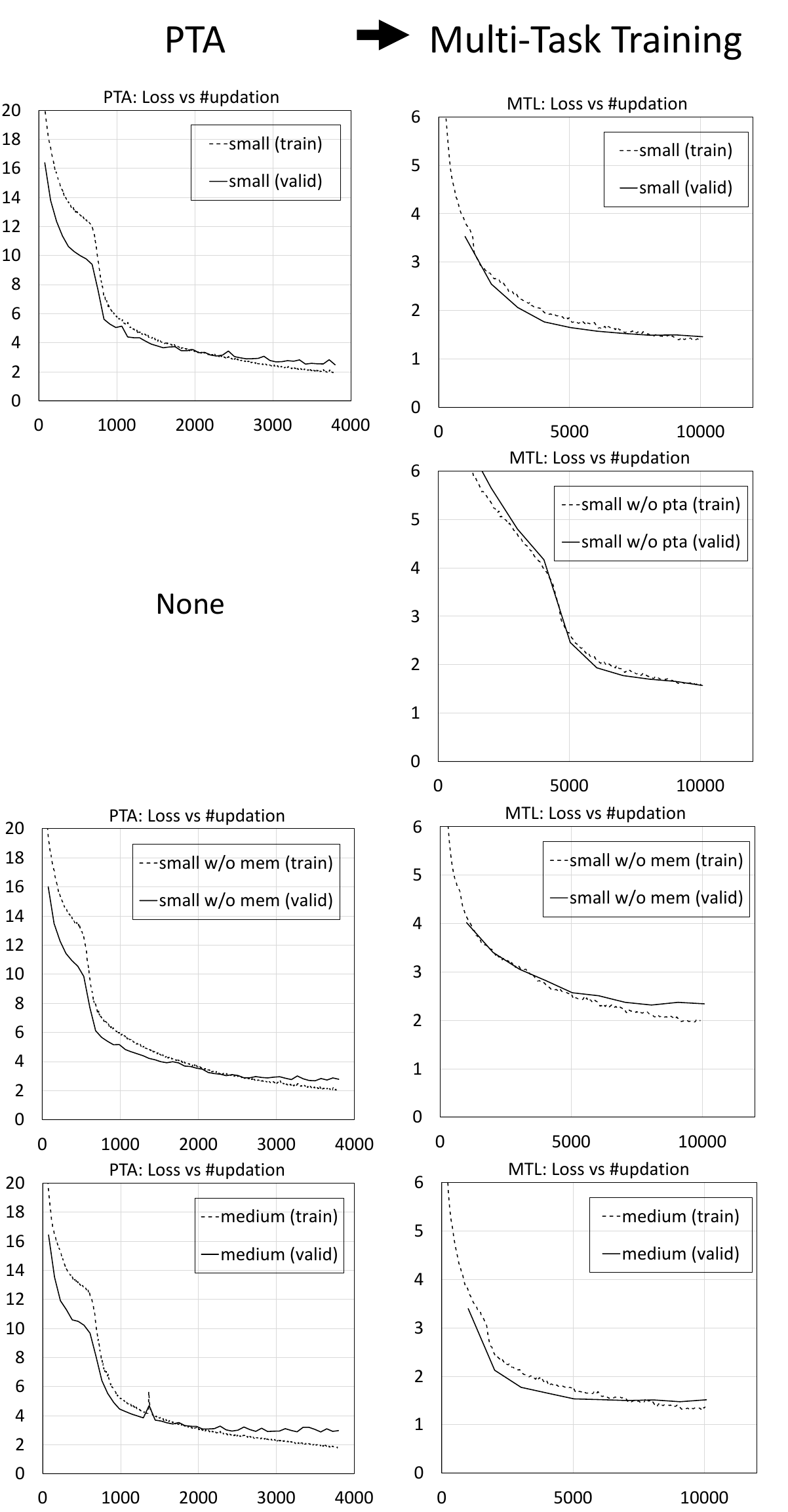}
\vspace{-2mm}
\caption{\label{figure:learning_curvers}
Learning curves of the BUI models.
}
\end{figure}

Figure~\ref{figure:learning_curvers} shows the learning curves of the BUI models.
In the PTA training, three models were roughly converged.
In the multi-task training, all models except BUI-BERT$_{\rm small}$ w/o PTA were roughly converged in 10 epoch.
However, the loss of BUI-BERT$_{\rm small}$ w/o PTA began to reduce drastically around 5k update and it could become smaller after 10 epoch.
This indicate that PTA speeds up the convergence of the loss at least, but it may not affect the final performance achieved after longer time.

\subsection{Cases of Task Execution}

\begin{figure*}
\centering
(a) Failure (timeout). (CoLA val. 554)
\includegraphics[width=\linewidth]{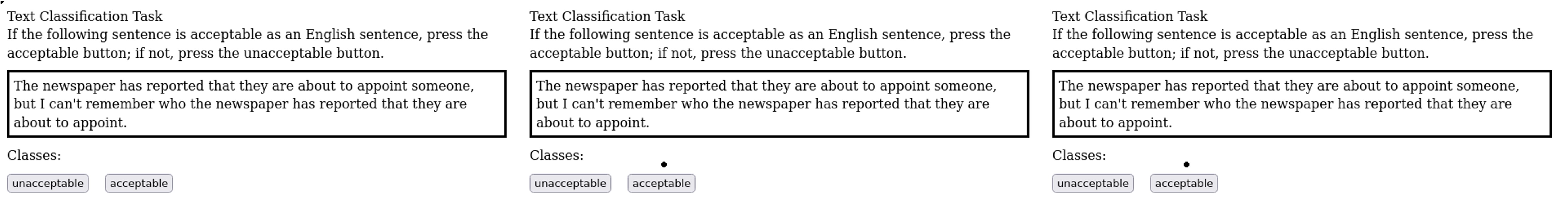}
(b) Success. (SA val. 34)
\includegraphics[width=\linewidth]{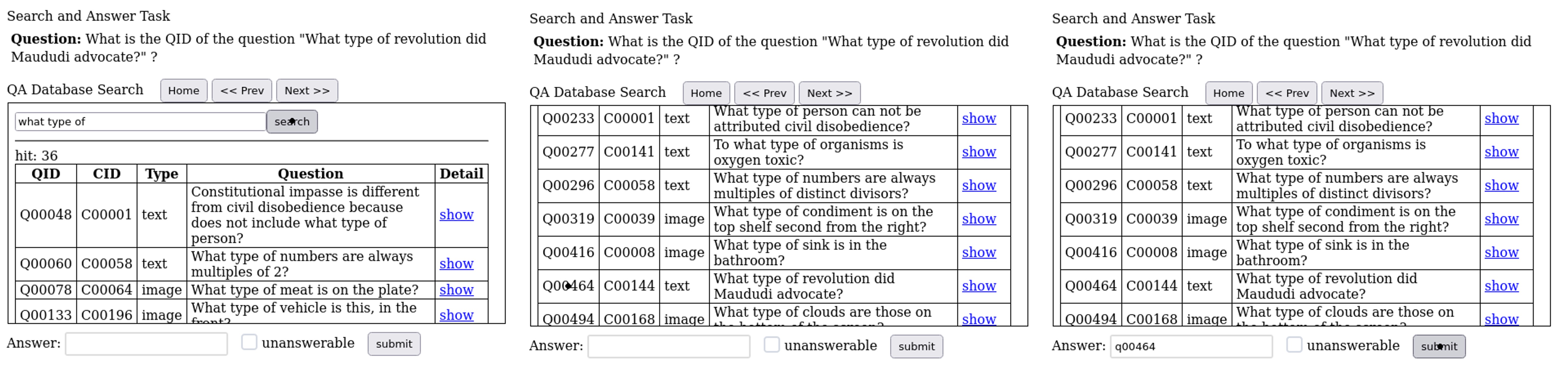}
(c) Failure. Gold answer : article 30, model : unanswerable (SA val. 42)
\includegraphics[width=\linewidth]{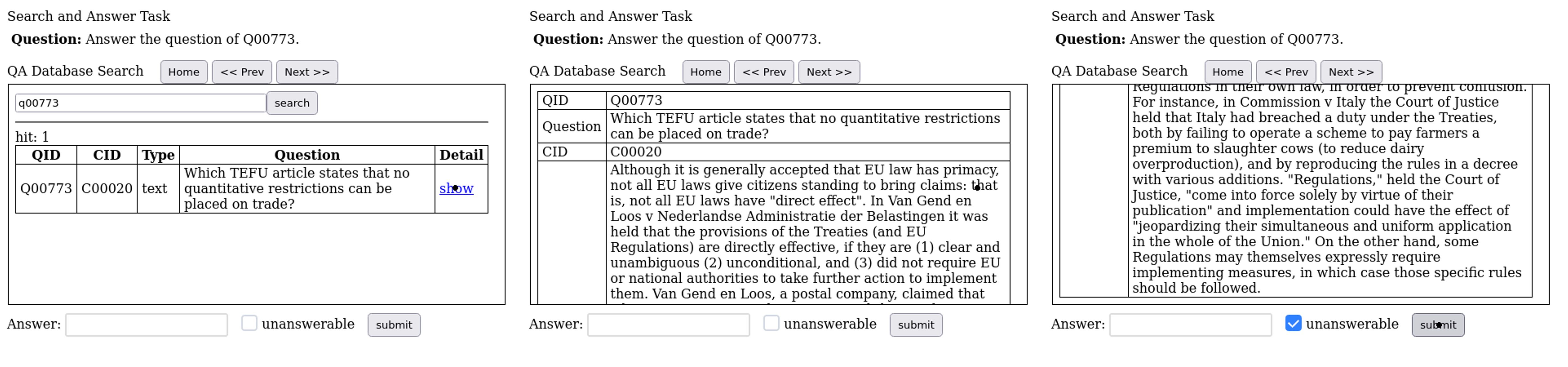}
(d) Failure. Gold answer : gray, model : blue. (SA val. 46)
\includegraphics[width=\linewidth]{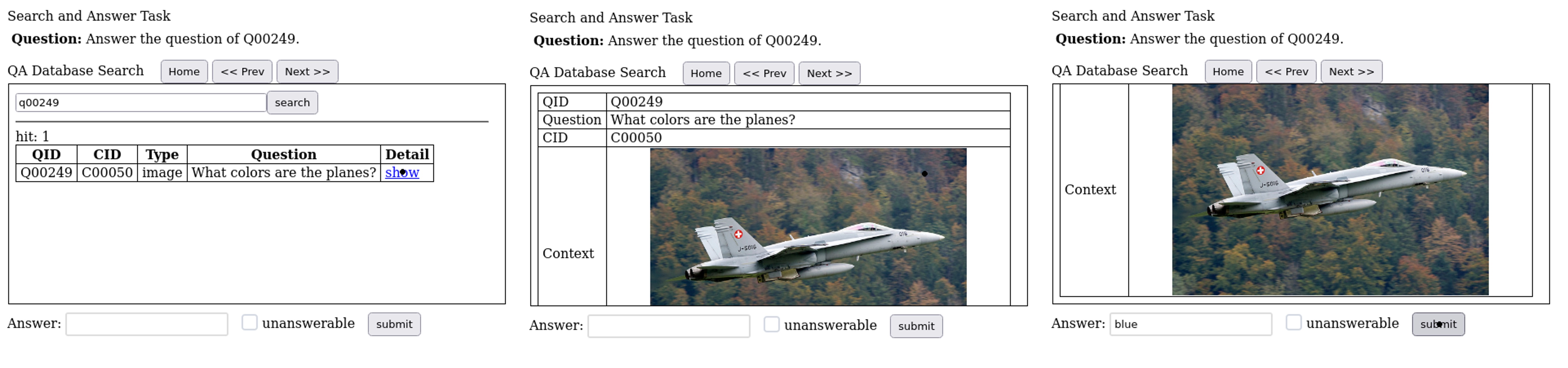}
(e) Success. Gold answer : third, model : third-most abundant element. (SA val. 67)
\includegraphics[width=\linewidth]{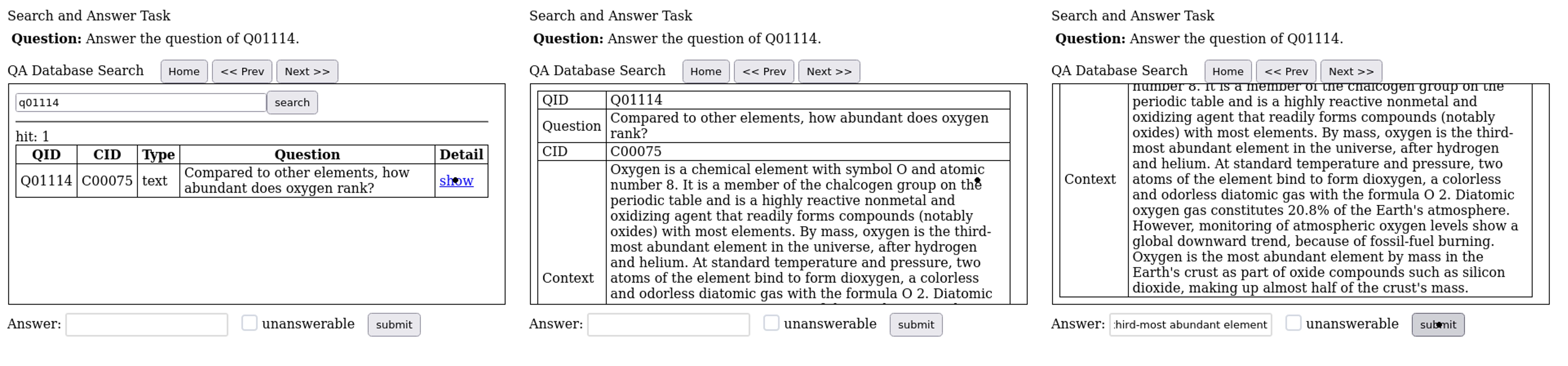}
\vspace{-2mm}
\caption{\label{figure:task_cases} Case studies. 
(a) Model repeated move\_to (172, 200), click, token (“unacceptable”), move\_to (172, 178), click, token (“unacceptable”), move\_to (172, 200), …
(b) Model queried the first three words, which is the same strategy as the gold sequence, and obtained a list. It scrolled down until the question appeared and then extracted the QID successfully.
(c) Model went to the detail and read all the context.
However, it chose the unanswerable check box to an answerable question.
(d) Model went to the detail to see the picture.
The answer type was correct, but the answer was different to the gold answer.
(e) Model went to the detail and read all the context to answer correctly.
}
\end{figure*}

We show the cases of task execution using BUI-BERT$_{\rm small}$ in Figure~\ref{figure:task_cases} as an aid to understanding.

\end{document}